\documentclass[acmtog, nonacm]{acmart}
\AtBeginDocument{%
  }

\setcopyright{acmcopyright}
\copyrightyear{2026}
\acmYear{2026}
\acmDOI{XXXXXXX.XXXXXXX}
\acmConference[Conference acronym 'XX]{Make sure to enter the correct
  conference title from your rights confirmation email}{June 03--05,
  2018}{Woodstock, NY}

\newcommand{\etal}{{\em{et~al.}\ }}

\newcommand{\vc}[1]{\ensuremath{\mathbf{#1}}}

\usepackage{algorithmic}
\usepackage{graphicx}
\usepackage[ruled]{algorithm2e} 

\SetAlFnt{\small}
\SetAlCapFnt{\small}
\SetAlCapNameFnt{\small}
\SetAlCapHSkip{0pt}
\usepackage{subcaption}
\usepackage[table]{xcolor}
\usepackage{multirow}

\begin{document}

\title{DSD: Learning Diverse and Reusable Motor Skills via Diffusion Skill Discovery}

\author{Sun Woo Kim}
\affiliation{%
  \institution{Simon Fraser University}
  \country{Canada} 
}
\email{ska503@sfu.ca}

\author{Xue Bin Peng}
\affiliation{%
  \institution{Simon Fraser University}
  \country{Canada} ,
  \institution{NVIDIA}
  \country{Canada}
}
\email{xbpeng@sfu.ca}

\renewcommand{\shortauthors}{Kim\etal}

\begin{abstract}
Humans efficiently learn new tasks by reusing a rich repertoire of motor skills across different goals and contexts. A similar strategy can also be used to enable simulated characters to efficiently perform new tasks by leveraging reusable motor skills.  
To support a wide range of downstream tasks, the learned repertoire should be diverse, consisting of distinct behaviors as well as spatial and temporal variation within each behavior. A commonly used method for learning diverse skills is by maximizing the mutual information between skill latents and the states produced by a policy. The marginal state entropy promotes broad behavioral coverage, while the conditional entropy encourages consistent behaviors from each latent. However, directly estimating the marginal state entropy is intractable in high-dimensional control problems. Prior methods therefore rely on indirect latent-space approximations or coarse estimators of the state distribution. These approximations may not effectively promote broad coverage of the state space, resulting in skills with limited behavioral diversity and reduced utility for downstream tasks. In this work, we propose Diffusion Skill Discovery (DSD), a skill discovery method that uses a diffusion model to approximate the entropy gradient of the policy-induced state distribution through score matching. The resulting objective encourages the discovery of skills that produce a broader range of behaviors for high-dimensional humanoid control. The learned skills are reused in two downstream control settings: hierarchical control with a task-specific high-level policy and zero-shot control through latent selection from offline trajectories. Our experiments show that DSD discovers a broader repertoire of reusable motor skills than prior skill discovery methods, leading to the emergence of complex and agile behaviors that can be reused across downstream tasks. Video available at https://youtu.be/QhMs67fvuWk.
\end{abstract}

\ccsdesc[500]{Computing methodologies~Procedural animation}
\ccsdesc[500]{Computing methodologies~Reinforcement learning}
\ccsdesc[300]{Computing methodologies~Unsupervised learning}
\keywords{character animation, reinforcement
learning, unsupervised reinforcement learning,  score distillation
sampling, diffusion model 
}

\begin{teaserfigure}
\includegraphics[width=\textwidth]{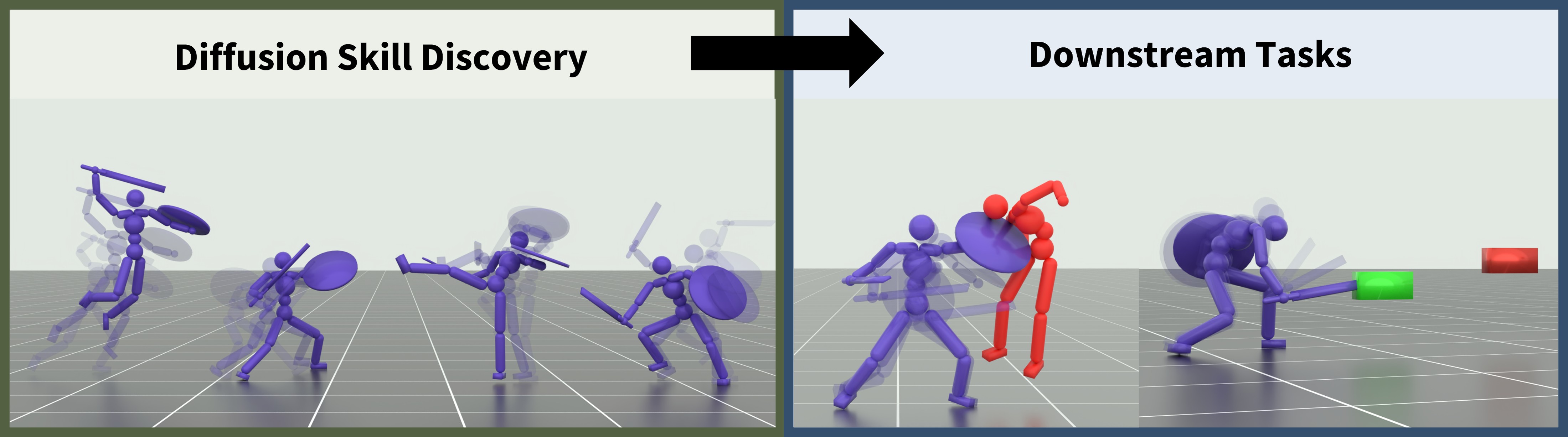}
\caption {Our method discovers diverse reusable motor skills through a diffusion-based  mutual information objective (left). The learned skills can then be reused to perform downstream tasks (right).}
\label{fig:teaser}
\end{teaserfigure}


\maketitle
\section{Introduction}
\label{sec:intro}
From everyday locomotion to complex athletic routines, human behavior is built on a rich repertoire of reusable motor skills. These skills allow humans to adapt familiar behaviors to new goals, contexts, and environments, avoiding the need to learn each task from scratch. For physically simulated humanoid characters, achieving flexible control beyond task-specific settings requires models that can develop and repurpose diverse motor skills, akin to those observed in human behaviors.

To endow simulated characters with diverse and reusable motor skills, a common approach is to model reusable skills as latent variable models. These methods combine a skill discovery objective, which encourages different latents to produce different behaviors, with an imitation objective, which keeps the behaviors close to a reference motion distribution, such that the discovered skills remain naturalistic~\citep{dou2022case,peng2022ASE,tessler2023calm, tirinzoni2025zero}. The resulting latent space provides a skill repertoire that the agent can then reuse for new tasks.

For a learned skill repertoire to support a wide range of downstream tasks, it should contain diverse and versatile skills that can be reused under different task objectives. 
A useful repertoire should include skills that offer diverse utility, such as walking, running, jumping, or striking when those behaviors are represented in the dataset. The repertoire should also capture spatial and temporal variations of those behaviors. A repertoire with limited diversity may map different latents to similar behaviors with only minor variations, reducing the range of downstream tasks that can be solved by the skills.

Existing latent-conditioned skill discovery methods encourage diversity by maximizing mutual information between the latents and the resulting motion distribution~\citep{dou2022case,peng2022ASE,tessler2023calm}. A key challenge is estimating the entropy of the marginal state distribution induced by the policy, which is generally intractable in high-dimensional humanoid control problems. Practical methods therefore rely on variational approximations that avoid directly modeling the policy-induced state distribution, and instead optimize surrogate objectives in the latent space~\citep{eysenbach2018diversity,gregor2016variational,baumli2021relative,hansen2020Fast}. While effective in simple
settings, such approximations can often lead to skills that exhibit only minor variations that offer little functional variability~\citep{campos20exp}.
More direct entropy estimators, such as particle-based estimators~\citep{liu2021aps} or surrogate objectives based on skill dynamics~\citep{sharma2019dynamics}, can provide coarse estimates of state diversity. However, nearest-neighbor distance estimates and skill-conditioned dynamics predictions can become unreliable in high-dimensional state spaces. As a result, the learned skills can lack functional versatility, thereby hampering their utility for downstream tasks.

To improve behavioral diversity for skill discovery with high-dimensional humanoid agents, we introduce Diffusion Skill Discovery (DSD). DSD jointly trains a diffusion model with a policy to approximate the gradient of the state-entropy through score matching. This approximation provides a practical surrogate for state-entropy maximization with high-dimensional systems. The resulting objective encourages the policy to visit a broader range of novel states, thereby increasing behavioral diversity across the learned skills. Once trained, the learned repertoire can be reused in downstream tasks through either task-specific high-level policies or zero-shot latent selection from an offline trajectory dataset.

The central contribution of this work is a diffusion-based skill discovery method that learns reusable motor skills from unstructured motion datasets. We introduce a number of practical design decisions for applying diffusion-based entropy maximization to high-dimensional humanoid control, improving training stability and motion quality. To our knowledge, DSD is the first unsupervised skill discovery framework for physically simulated humanoids to approximate the state-entropy gradient with a diffusion model. Experiments across multiple motion datasets show that DSD learns a more diverse repertoire of motion skills than prior skill discovery methods, including complex and agile motions. The utility of the learned repertoire is demonstrated in challenging downstream tasks using both hierarchical control and zero-shot latent selection.

\section{Related Works}
\label{sec:related}

Developing motor controllers for physically simulated characters has long been a central challenge in character animation. Early efforts in this domain leveraged optimization-based methods with hand-designed objectives to create controllers capable of producing natural motions~\cite{hodgins1995animating, coros2010Generalized, raibert1991animation, yin2007simbicon, zordan2002mocap}. By optimizing task-specific cost functions and constraints, these methods can produce high-quality motions for specific behaviors~\cite{Panne1994virtual, geyer2003positive, deLasa2010Feature, levine132013guided, mordatch2012cio, tan2014bicycle, ha2012falling}. However, these methods often require careful design of the objective and optimization process for each skill, which ultimately limits their effectiveness when applied to train controllers for a large corpus of skills~\cite{gejitenbeek2013flexible, wang2012optimizing, lee2009biomech}.

\subsection{Data-Driven Character Control}
To reduce the manual effort involved in designing controllers and objective functions that produce natural motions, data-driven approaches that leverage motion data have become widely adopted in physics-based character animation~\cite{da2008simulation, sharon2005synthesis, kwon2017momentum, lee2010biped, zordan2002mocap}. Early data-driven methods often depended on carefully engineered, skill-specific controllers to track reference motion clips~\cite{liu2005learning, liu2010sampling, liu2012terrain, sok200simulating, liu2016guided}. Deep reinforcement learning later provided a more general framework for motion imitation by replacing these specialized control structures with neural network policies trained through a reward-driven optimization framework~\cite{peng2018deepmimic, chentanez2018physics, lee2019scalable}.
While reinforcement learning presents a versatile framework for training control policies, learning a single policy that covers a broad range of motions remains challenging. Prior work has addressed this challenge by incorporating motion planners that coordinate multiple controllers to cover a broader set of behaviors~\cite{coros2009robust, hausman2018learning, heess2016learning}. Another line of work uses adversarial imitation objectives to train a single policy to match the distribution of reference motion data~\cite{peng2021AMP, zhang2025ADD}. This connects to motion coverage because the discriminator compares the policy against the full reference distribution rather than a single target motion. Our method uses this adversarial formulation as a motion prior to guide the policy toward naturalistic behaviors, while the diffusion-based skill discovery objective promotes broad behavioral coverage and organizes the resulting behaviors into distinct latent-variable skills.

\subsection{Unsupervised Skill Discovery}
Unsupervised reinforcement learning aims to discover skills without relying on extrinsic task rewards. Policies learn diverse behaviors by optimizing intrinsic objectives, which enables the model to automatically discover reusable skills that can be repurposed for downstream tasks.
Common intrinsic objectives rely on prediction errors from learned models~\cite{pathak2017exploration, burda2018exploration}, visitation-based counts~\cite{bellemare2016unifying, tang2017exploration}, or entropy maximization that promote broad exploration of the state space~\cite{hazan2019provably, liu2021behavior, ying2025exploratorydiffusion}. These objectives promote state-space exploration, but they do not necessarily produce a reusable skill repertoire. Latent-conditioned methods instead train policies whose behaviors depend on auxiliary latents, using approximate mutual-information objectives~\citep{gregor2016variational,eysenbach2018diversity,liu2021aps} or alternative Lipschitz-constrained measures~\citep{park2021lsd,park2024metra} to encourage different latents to produce different behavior. Another line of work trains policies that can be prompted by rewards or goals without additional policy optimization~\citep{touati2021learning,frans2024unsupervised}. These methods support flexible downstream adaptation, but typically require separate exploration procedures to produce behavioral diversity. Recent work combines an intrinsic skill discovery objective with a motion imitation objective to obtain skills that produce naturalistic behaviors~\cite{peng2022ASE, tessler2023calm, dou2022case, tirinzoni2025zero}. Although existing methods learn reusable motor skills, discovering a broadly diverse repertoire remains challenging. DSD addresses this limitation with a diffusion-based approximation that promotes broader coverage of the policy's state distribution.

\subsection{Diffusion-based Distribution Modeling}
Diffusion models have demonstrated strong performance as generative models across many domains, including human motion, by modeling data distributions through iterative denoising~\cite{cohan2024flexible, kim2023FLAME, tevet2023human}. Beyond sample generation, score distillation sampling (SDS) uses a pre-trained diffusion model to estimate gradients that guide an optimization process towards samples that maximize the likelihood under a target data distribution~\cite{poole2023dreamfusion,jiang2024animated,wang2023prolific,yin2024onestep}. Diffusion models have also been used for motion imitation in reinforcement learning, where a pretrained diffusion model guides a policy toward motions that resemble a reference motion distribution~\cite{luo2024text,wu2024diffusing,mu2025smpreusablescorematchingmotion}. In addition to training policies to imitate a target behavioral distribution, diffusion models have also been used as an exploration objective to encourage diverse behaviors. Most closely related to our work, ExDM uses diffusion-based score estimation to approximate a marginal state-entropy objective for unsupervised exploration~\citep{ying2025exploratorydiffusion}. However, ExDM is designed solely for exploration and does not learn a reusable latent representation of skills. DSD instead incorporates skill latents within a mutual-information objective to learn a reusable latent representation for downstream tasks. Furthermore, ExDM maximizes state diversity without considering motion naturalness, which can result in unrealistic behaviors. To address this, DSD leverages a motion prior to guide skill discovery toward natural human motions. Finally, ExDM was demonstrated only in relatively simple and low-dimensional domains. In this work, we introduce practical design decisions for effectively applying diffusion-based entropy estimation to high-dimensional humanoid control.

\section{Background}
\label{sec:background}
In this section, we provide a review of the core machine learning concepts behind our framework: reinforcement learning and diffusion model formulations. 

\subsection{Reinforcement Learning}
Our framework is modeled using reinforcement learning, where an agent interacts with its environment according to a policy $\pi$ in order to maximize a given objective~\cite{sutton1998rl}. For simplicity, we approximate the formulation as a Markov decision process (MDP). At each time step, the agent observes the state $\vc{s}_t$ and samples an action $\vc{a}_t$ from the learned policy $\pi(\vc{a}_t | \vc{s}_t)$. The next state $\vc{s}_{t+1}$ is determined by the environment’s transition dynamics $p(\vc{s}_{t+1} | \vc{s}_t, \vc{a}_t)$. After each transition, the agent receives a scalar reward $r_t$.
The objective of the agent is to maximize the discounted cumulative return 
\begin{equation}
    J = \mathbb{E}_{p(\tau| \pi)} \left [  \sum_{t=0}^T \gamma^t r_t \right ],
\end{equation}
where a trajectory $\tau$ is the sequence of states and actions over a time horizon, $\tau = (\vc{s}_0, \vc{a}_0, \vc{s}_1, \vc{a}_1, \ldots , \vc{s}_{T-1}, \vc{a}_{T-1}, \vc{s}_T, \vc{a}_T)$ and $\gamma$ is a discount factor. Where as the likelihood of a trajectory under the policy $p(\tau|\pi)$ is represented as $p(\tau | \pi) = p(\vc{s}_0)\Pi_{t=0}^{T-1}p(\vc{s}_{t+1} | \vc{s}_t, \vc{a}_t) \pi(\vc{a}_t | \vc{s}_t)$.

\subsection{Diffusion Model}
Diffusion models learn a data distribution by reversing a gradual noising process. Given a clean sample $\vc{x}^0$, the forward process constructs a noisy sample $\vc{x}^k$ at diffusion step $k$ as
\begin{equation}
    \vc{x}^k = \sqrt{\bar{\alpha}_k}\,\vc{x}^0 + \sqrt{1-\bar{\alpha}_k}\,\boldsymbol{\epsilon}, 
\quad \boldsymbol{\epsilon}\sim \mathcal{N}\left(0, I \right ),
\label{eq:diffusion}
\end{equation}
where $\boldsymbol{\epsilon}$ is Gaussian noise, $\bar{\alpha}_k$ denotes the cumulative product of a fixed noise schedule, and $k \sim p(k)$ denotes the diffusion timestep. The model is then trained to reverse this forward process by iteratively denoising from Gaussian noise back toward the data distribution with a DDPM (Denoising Diffusion Probabilistic Models) objective~\citep{ho2020ddpm}:
\begin{equation}
\min_{\theta} 
\mathbb{E}_{k, \vc{x}^0, \boldsymbol{\epsilon}} w\left ( k \right ) \left\lVert  \boldsymbol{\epsilon}_{\theta}\left ( \vc{x}^k \right ) - \boldsymbol{\epsilon} \right\rVert^2 ,
    \end{equation}
where $w(k)$ denotes the weight associated with diffusion step $k$.

Beyond sample generation, a diffusion model can also be repurposed into an optimization objective through score distillation sampling (SDS)~\cite{poole2023dreamfusion}. In SDS, a pretrained diffusion model provides score-based gradients that optimize the current parameterized sample distribution toward the data distribution encoded by the model. A commonly used SDS surrogate is
\begin{equation}
\min_{\vc{x}^0}
\mathbb{E}_{k, \boldsymbol{\epsilon}}
\left[
w\left ( k \right )\left\lVert \boldsymbol{\epsilon}_{\theta} \left (\vc{x}^k \right ) - \boldsymbol{\epsilon} \right\rVert^2
\right].
\end{equation}
This objective leads to gradients that increase the likelihood of samples under the pretrained diffusion model, thereby acting as a distribution-matching objective to produce samples that resemble the original dataset.

\begin{figure}[t]
\centering
\includegraphics[width=0.95 \columnwidth]{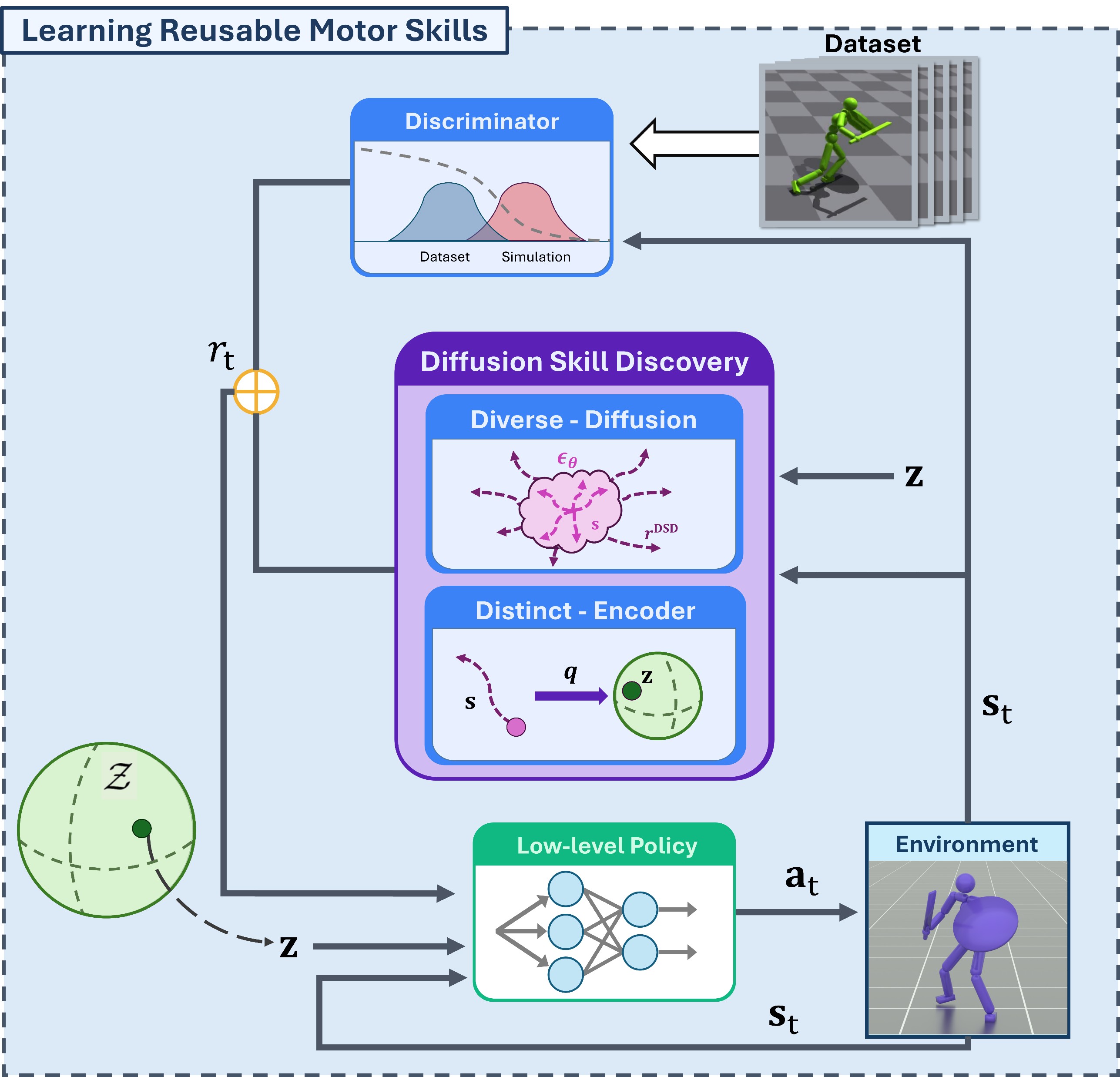}
\caption{Overview of our DSD framework for learning reusable motor skills. The low-level policy is conditioned on a skill latent $\vc{z}$, which generates trajectories through environment interactions. DSD promotes diversity using a diffusion model $\epsilon_\theta$ and distinct skills through a skill encoder $q$. A GAN-style discriminator $D$ encourages natural motions, and the combined reward $r_t$ is then used to train the policy.}
\Description{DSD low level policy diagram}
\label{fig:systemdiagram1}
\end{figure}

\begin{figure}[h]
\centering
\includegraphics[width=0.95 \columnwidth]{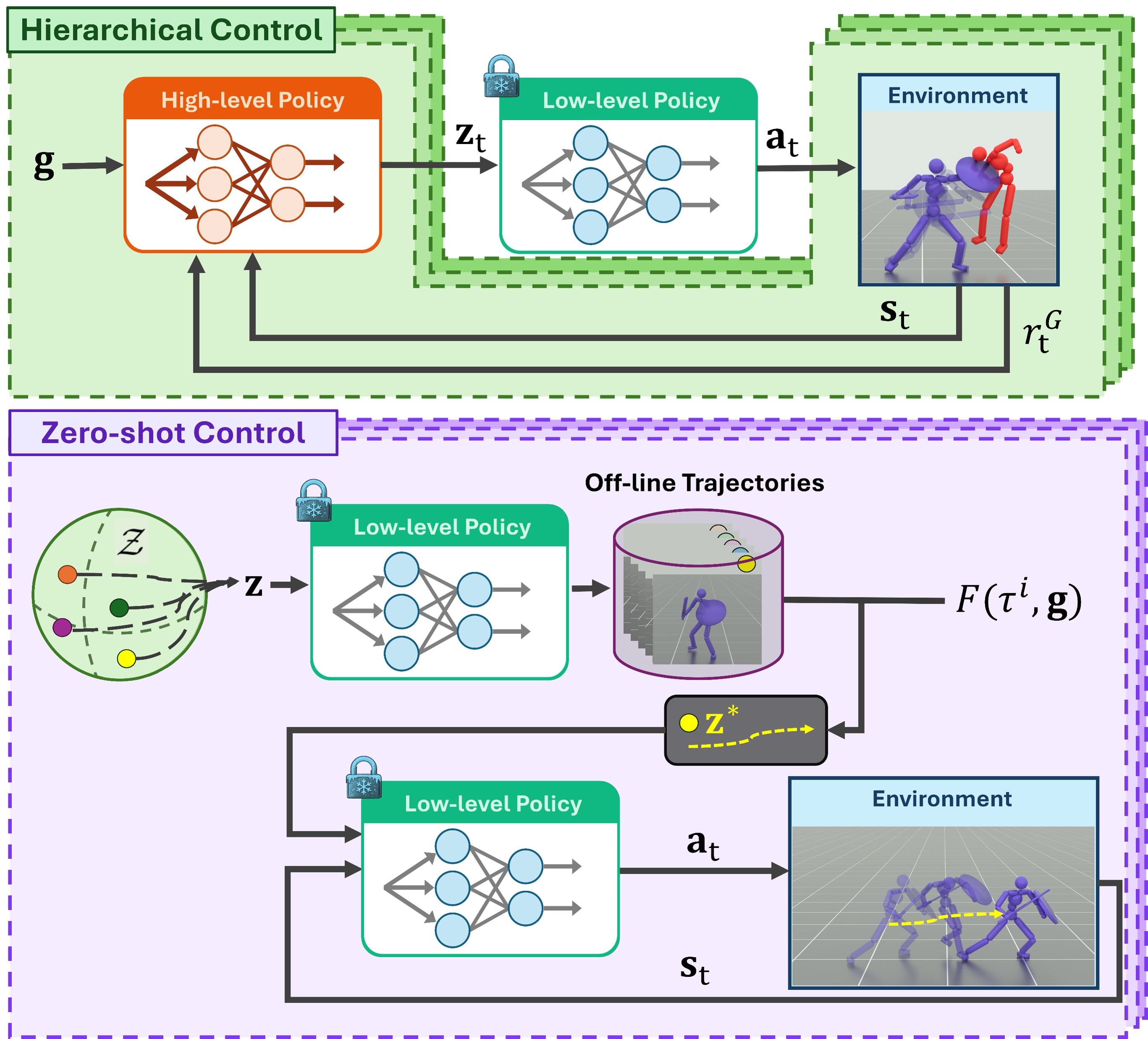}
\caption{Once the low-level policy has been trained, its skills can be reused to perform new downstream tasks through two different methods. Top: Hierarchical control, where a high-level policy outputs latents for directing the frozen low-level policy. Bottom: Zero-shot control, where candidate latents are evaluated using an offline trajectory dataset, and the latent $\vc{z}^{*}$ that maximizes the fitness function $F$ is selected to drive the frozen low-level policy, without requiring additional training.}
\Description{DSD downstream task}
\label{fig:systemdiagram2}
\end{figure}

\section{Overview}
\label{sec:overview}
In this work, we introduce Diffusion Skill Discovery (DSD), a framework that discovers diverse and reusable motor skills from unlabeled motion data through a diffusion-based skill discovery objective. While SDS methods use pretrained diffusion models to guide an optimization process towards a target distribution, DSD inverts this idea and uses diffusion models to encourage exploration and maximize diversity. As illustrated in Figure~\ref{fig:systemdiagram1}, DSD trains a generative policy $\pi\left (\vc{a}|\vc{s},\vc{z} \right )$ where the skill latent $\vc{z}$ specifies different skills. The diffusion model $\boldsymbol{\epsilon}_{\theta}$ is jointly trained with the policy on the states visited during policy rollouts. The diffusion model estimates the entropy gradient of the policy-induced state distribution, which provides a reward that encourages the policy to visit novel states. This objective promotes behavioral diversity across the learned skills. A latent-conditioned encoder $q\left(\vc{z}|\vc{s}\right)$ encourages different latents to produce distinct skills. An adversarial imitation objective uses a reference motion dataset to keep the discovered behaviors naturalistic. Together, these objectives produce a diverse skill repertoire that reflects broad behavioral variations represented in the motion data.

Once trained, the low-level policy can be reused for downstream tasks, as illustrated in Figure~\ref{fig:systemdiagram2}. The first reuse strategy trains a task-specific high-level policy $\omega\left(\vc{z}|\vc{s},\vc{g}\right)$, where $\vc{g}$ is a task-specific goal. Following prior hierarchical control frameworks~\citep{peng2022ASE}, the high-level policy selects latents that direct the low-level policy toward the goal. The second strategy performs zero-shot latent selection without additional model training. Zero-shot control selects a skill latent for the low-level policy based on task-specific fitness evaluated over an offline trajectory pool.

\begin{figure*}[t]
\centering
\includegraphics[width=0.95\linewidth]{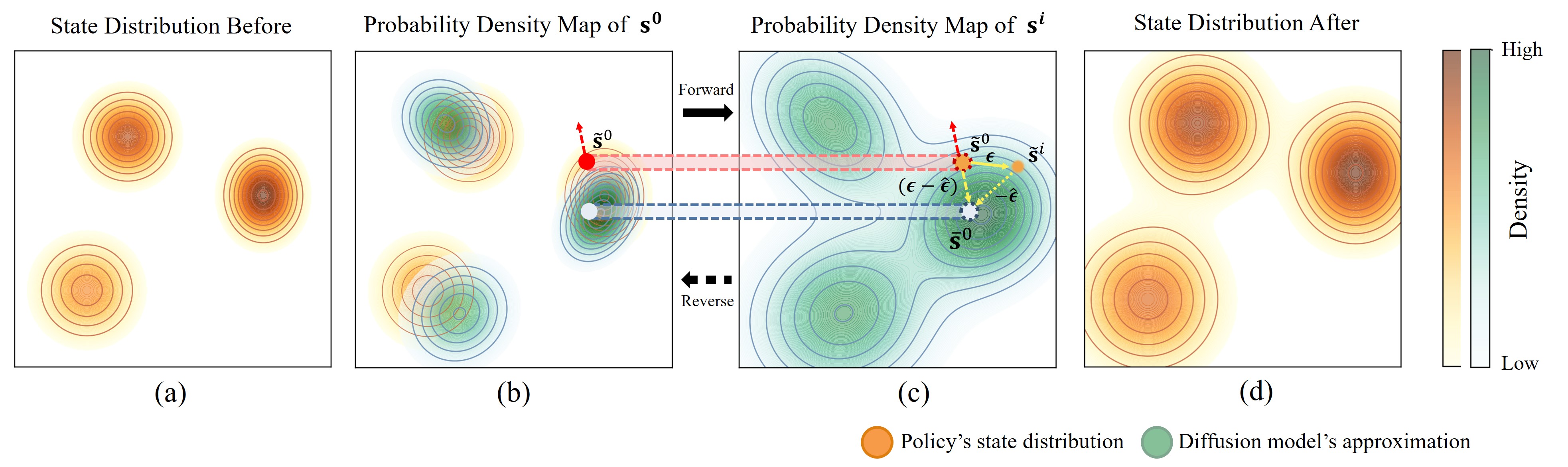}
\caption{An illustration of entropy maximization through diffusion-based score estimation.  At each iteration, the policy induces a state distribution, shown in orange (a).
A diffusion model is trained on samples from the policy shown in green (b). Through the forward diffusion process, the diffusion model predicts the added noise $\hat{\boldsymbol{\epsilon}}$. The residual $(\boldsymbol{\epsilon} - \hat{\boldsymbol{\epsilon}})$ determines an update direction toward a pseudo target $\bar{\mathbf{s}}^0$ that increases the likelihood under the state distribution (c). Conversely, maximizing the prediction error encourages the policy to visit novel states with lower likelihoods under the current policy, thereby expanding the state distribution and producing more diverse behaviors at the next iteration (d).}
\label{fig:visual_exp}
\end{figure*}

\section{Diffusion Skill Discovery}
\label{sec:dsd}
DSD discovers diverse and distinct skills through mutual information maximization. Let $S$ denote the random variable over policy-induced states and $Z$ denote the skill latent. The objective is derived from the mutual information between $S$ and $Z$:
\begin{equation}
I(S;Z) = \mathcal{H}(S) - \mathcal{H}(S | Z).
\end{equation}
The marginal entropy $\mathcal{H}(S)$ encourages broad coverage of the policy-induced state distribution, while minimizing the conditional entropy $\mathcal{H}(S | Z)$ encourages different latents to induce distinct behaviors. Since estimating both state distributions is generally intractable, DSD approximates the marginal entropy using a diffusion model and the conditional entropy using a skill encoder.

\subsection{Marginal Entropy Maximization}
\label{sec:marginal_entropy}
The marginal entropy of a policy's state distribution is given by 
\begin{equation}
\mathcal{H}(S) = \mathbb{E}_{d^{\pi}(\vc{s})} \left[ -\log p(\vc{s}) \right],
\end{equation} where $d^{\pi}(\vc{s})$ denotes the marginal state distribution induced by a policy $\pi(\vc{a} | \vc{s}, \vc{z})$.
Estimating the marginal entropy requires evaluating the log-likelihood of states under the policy, which is generally not readily available. To construct a tractable estimate of the negative log-likelihood, we propose approximating the policy's marginal state distribution using a diffusion model trained on samples collected from the policy's rollouts. Given states observed from the policy rollouts, a diffusion model can be trained using the standard DDPM loss~\cite{ho2020ddpm}. Once trained, the diffusion model can be used to specify an exploration reward through score-distillation sampling~\cite{poole2023dreamfusion}, which provides a gradient for increasing the entropy of the policy's state distribution.

The diffusion model can be trained with the DDPM objective:
\begin{equation}
\min_{\theta} \quad
\mathbb{E}_{d^{\pi}(\vc{s}),\, k \sim \mathcal{U}(0, N)} \left [w\left ( k \right )
\left\lVert
\boldsymbol{\epsilon}_{\theta}\left( \vc{s}^k \right)
-
\boldsymbol{\epsilon}
\right\rVert^2  \right],
\label{eq:diffusion_obj}
\end{equation}
where $k$ is a randomly sampled diffusion timestep, $\vc{s}^k$ is the noisy state at timestep $k$, $\boldsymbol{\epsilon}$ is the added noise, and the denoising model $\boldsymbol{\epsilon}_{\theta}$ is trained to predict the added noise. The timestep weights $w(k)$ are determined by a noise schedule $\bar{\alpha}$. The weights are commonly omitted to simplify the DDPM objective~\cite{ho2020ddpm}. The DDPM objective is related to a variational upper bound on the negative log-likelihood of samples generated by the diffusion model:
\begin{equation}
\mathbb{E}_{\vc{s} \sim d^{\pi}}
\left[
-\log p(\vc{s})
\right]
\leq
\mathbb{E}_{\vc{s} \sim d^{\pi},\, k \sim \mathcal{U}(0, N)}
\left[
w(k)
\left\lVert
\boldsymbol{\epsilon}_{\theta}\left( \vc{s}^k \right)
-
\boldsymbol{\epsilon}
\right\rVert^2
\right]
+ C,
\label{eq:ddpm_upper}
\end{equation}
where $C$ is a constant. This relationship connects diffusion reconstruction error to density modeling. We leverage this connection to construct an exploration objective by using the diffusion reconstruction error to estimate gradients for increasing the entropy of a policy's marginal state distribution.

To train the policy, the diffusion model is used to evaluate the reconstruction loss on the policy's observed states. This leads to the following marginal entropy reward:
\begin{equation}
r_t^{\mathrm{ME}}
=
\sum_{k \in \mathbb{K}}
\left\lVert
\boldsymbol{\epsilon}_{\theta}\left( \vc{s}_t^k \right)
-
\boldsymbol{\epsilon}
\right\rVert^2 ,
\label{eq:marginal}
\end{equation}
where $\mathbb{K}$ denotes a set of diffusion timesteps used for reward evaluation.  Maximizing this reward pushes the policy-induced state distribution toward rare states that are less accurately modeled by the diffusion model. The diffusion model is then updated using the policy's observed states, thereby continuously encouraging the policy to visit novel states. Figure~\ref{fig:visual_exp} illustrates this update process.

Since Equation~\ref{eq:ddpm_upper} is an upper bound on $\mathcal{H}(S)$, maximizing it does not ensure an increase in the underlying marginal entropy $\mathcal{H}(S)$. However, empirically we show that this objective leads to an effective reward function for diversity maximization. This objective is similar to prior exploration methods that utilize prediction errors as a novelty-seeking objective~\cite{burda2018exploration}. Since the diffusion model is trained on data from the policy, the prediction errors should be low under states similar to those previously visited by the policy. Conversely, novel states that differ from commonly visited states should lead to larger prediction errors from the diffusion model. 

\subsection{Conditional Entropy Minimization}
\label{sec:conditional_entropy}
The conditional entropy $\mathcal{H}(S | Z)$ requires modeling the conditional state distribution for each skill latent $\vc{z}$. For each $\vc{z}$, let $d^\pi(\vc{s} | \vc{z})$ denote the distribution over states induced by $\pi(\vc{a} | \vc{s}, \vc{z})$, and let $p(\vc{s} | \vc{z})$ denote the likelihood of a state $\vc{s}$ given a skill $\vc{z}$. DSD utilizes a variational approximation $q(\vc{s} | \vc{z})$ to obtain a lower bound on $-\mathcal{H}(S | Z)$:
\begin{equation}
\begin{aligned}
-\mathcal{H}(S | Z)
&=
\mathbb{E}_{p(\vc{z})}
\mathbb{E}_{d^{\pi}(\vc{s} | \vc{z})}
\left[
\log p(\vc{s} | \vc{z})
\right] \\
&\geq
\mathbb{E}_{p(\vc{z})}
\mathbb{E}_{d^{\pi}(\vc{s} | \vc{z})}
\left[
\log q(\vc{s} | \vc{z})
\right].
\end{aligned}
\end{equation}
Directly modeling the conditional distribution over states $q(\vc{s} | \vc{z})$ can be impractical, as the normalization constant for each latent $\vc{z}$ requires integration over the high-dimensional state space. DSD instead parameterizes the conditional log-likelihood using a von Mises--Fisher distribution. The latent space is defined as the surface of a unit hypersphere
$\mathcal{Z} = \{\vc{z} : \lVert \vc{z} \rVert = 1\}$.
The skill encoder maps each state to a unit vector $\mu_q(\vc{s}) \in \mathcal{Z}$, following the normalized skill-encoder parameterization used in ASE~\citep{peng2022ASE}. Following \citet{liu2021aps}, DSD uses the symmetry of the von Mises--Fisher distribution between the state embedding $\mu_q(\vc{s})$ and the latent $\vc{z}$ to approximate $\log q(\vc{s} | \vc{z})$, without explicitly modeling the normalized likelihood over states. This leads to the following conditional log-likelihood approximation:
\begin{equation}
\log q(\vc{s} | \vc{z})
\propto 
\kappa \, \mu_q(\vc{s})^T \vc{z},
\end{equation}
where $\kappa$ is a manually-specified scaling factor. The encoder is trained to predict the latent $\vc{z}$ that led to a particular state $\vc{s}$:
\begin{equation}
\max_q \quad
\mathbb{E}_{p(\vc{z})}
\mathbb{E}_{d^{\pi}(\vc{s} | \vc{z})}
\left[
\kappa \, \mu_q(\vc{s})^T \vc{z}
\right].
\label{eq:skill_enc_obj}
\end{equation}
The conditional entropy reward at each timestep $t$ for training the policy is then calculated according to:
\begin{equation}
r_t^{\mathrm{CE}}
= \kappa \, \mu_q(\vc{s}_t)^T \vc{z}_t.
\label{eq:conditional}
\end{equation}
Maximizing this reward incentivizes the policy $\pi$ to visit states where the encoder is able to accurately predict the skill latent $\vc{z}$. This in turn encourages the policy to produce a distinct state distribution for each skill such that the encoder can more accurately predict the corresponding $\vc{z}$.

\subsection{DSD Reward}

The full DSD reward combines the marginal and conditional entropy rewards:
\begin{equation}
r_t^{\mathrm{DSD}}
=
w^{\mathrm{ME}} r_t^{\mathrm{ME}}
+
w^{\mathrm{CE}} r_t^{\mathrm{CE}},
\label{eq:DSD_reward}
\end{equation}
with their respective weights $w^{\mathrm{ME}}$ and $w^{\mathrm{CE}}$. This combined reward is not a valid lower or upper bound on $I(S;Z)$, since $r_t^{\mathrm{ME}}$ is derived from a diffusion objective related to an upper bound on the marginal negative log-likelihood, whereas $r_t^{\mathrm{CE}}$ is derived from a variational lower bound on $-\mathcal{H}(S|Z)$. Instead, the sum of the two rewards acts as a practical surrogate for discovering diverse and distinct skills. 

\subsection{Implementation Decisions}
\label{sec:Practical}
Several implementation decisions are important for applying DSD effectively to humanoid control. These decisions address temporal coherency of skills, effective diffusion parametrization, overfitting of the diffusion model, and variance in entropy-reward estimation.  

\paragraph{State-History.}
The mutual information reward should capture behaviors that span multiple timesteps. To provide this temporal context, both the diffusion model and the skill encoder take as input an $N$-step state history $\vc{s}_{t-N+1:t}$ instead of a single state $\vc{s}_t$. A state-history length of $N = 10$ is used in all experiments.

\paragraph{Data-Prediction.}
DSD adopts the data-prediction parameterization commonly used in human motion diffusion models~\citep{tevet2023human,cohan2024flexible}, instead of the noise-prediction formulation originally proposed in DDPM~\citep{ho2020ddpm}. The diffusion model therefore directly predicts the denoised state $\hat{\vc{s}}^0$ at each diffusion step. Since the entropy reward in Eq.~\ref{eq:marginal} is defined using noise-prediction error, $\hat{\vc{s}}^0$ is converted into the corresponding noise estimate $\boldsymbol{\epsilon}_{\theta}$ for reward evaluation:
\begin{equation}
\boldsymbol{\epsilon}_{\theta}\left (\vc{s}^k \right ) = 
\frac{\vc{s}^k - \sqrt{\bar{\alpha}_k}\hat{\vc{s}}^0}
{\sqrt{1-\bar{\alpha}_k}}.
\label{eq:epsilon_theta}
\end{equation}
This conversion is used only for reward evaluation. The diffusion model is trained with the data-prediction parameterization.

\paragraph{Replay Buffer.}
Training the diffusion model only on trajectories generated by the policy at the current iteration can cause overfitting to the behavior of the most recent policy. To mitigate this overfitting, DSD trains the diffusion model using a replay buffer $\mathcal{R}$ containing samples collected across multiple previous iterations. At each iteration, newly collected trajectories are subsampled at a rate of $1/\eta$ and added to a replay buffer, where $\eta=32$.

\paragraph{Diffusion Timesteps.}
Randomly sampling the diffusion timesteps during reward evaluation can introduce high variance in the reward values, which can lead to less stable and slower training. To reduce this variance, DSD evaluates the reward at a fixed set of timesteps $\mathbb{K}$, following \citet{mu2025smpreusablescorematchingmotion}. The timesteps are set to $\mathbb{K} = {8, 15, 22}$ in all experiments.

\section{Motion Prior}
\label{sec:motionprior}
The skill discovery rewards encourage the policy to discover diverse and distinct skills, but they do not by themselves ensure natural human motions. DSD therefore incorporates human motion data through an adversarial motion prior (AMP), which guides the skill discovery process towards natural human-like behaviors. The AMP discriminator is trained on a reference motion dataset $\mathcal{M}=\left\{m^i\right\}$, where each $m^i$ is a motion clip captured from human actors. This setup enables direct comparison with ASE~\citep{peng2022ASE}, which also uses AMP reward as an imitation objective, while isolating the effects of the DSD reward. The motion prior is not required by the core DSD objective and can be replaced by other types of motion priors, such as C$\cdot$ASE~\citep{dou2022case} or SMP~\citep{mu2025smpreusablescorematchingmotion}.

The motion prior uses a discriminator $D$ to distinguish transitions from the reference motion distribution $d^{\mathcal{M}}\left(\vc{s},\vc{s}'\right)$ and the policy distribution $d^{\pi}\left(\vc{s},\vc{s}'\right)$. For clarity, the objective is written using a single transition pair, although in practice, $D$ receives the same $N$-step state history as the diffusion model and skill encoder. Following \citet{peng2021AMP}, the discriminator is trained with the objective:
\begin{equation}
\begin{aligned}
\min_{D} \quad
&- \mathbb{E}_{(\vc{s}, \vc{s}') \sim d^{\mathcal{M}}} \left[ \log D(\vc{s}, \vc{s}') \right]
- \mathbb{E}_{(\vc{s}, \vc{s}') \sim d^{\pi}} \left[ \log \left( 1 - D(\vc{s}, \vc{s}') \right) \right] \\
&\quad + w^{\mathrm{GP}} \, \mathbb{E}_{(\vc{s}, \vc{s}') \sim d^{\mathcal{M}}}
\left[
\left\lVert
\nabla_{(\vc{s}, \vc{s}')} D(\vc{s}, \vc{s}')
\right\rVert^2
\right],
\end{aligned}
\label{eq:fin_disc_obj}
\end{equation}
where $w^{\mathrm{GP}}$ is the weight for a gradient-penalty regularizer. The motion-prior reward is then given by
\begin{equation}
r_t^{\mathrm{AMP}} = -\log\left(1 - D(\vc{s}, \vc{s}')\right).
\label{eq:disc_reward}
\end{equation} 
This objective encourages $D$ to assign higher scores to reference motions from the dataset, and lower scores to motions from the policy. Together with the DSD reward, the full training reward for the low-level policy is defined by
\begin{equation}
r_t = w^{\mathrm{DSD}} r_t^{\mathrm{DSD}} + w^{\mathrm{AMP}} r_t^{\mathrm{AMP}},
\label{eq:full_reward}
\end{equation}
where $w^{\mathrm{DSD}}$ and $w^{\mathrm{AMP}}$ denote the weights for the DSD reward and AMP reward respectively. The combined reward function encourages the low-level policy to discover diverse and distinct skills that resemble the natural life-like behaviors depicted in the dataset.

\section{Jitter-Induced Diversity Suppression}
\label{sec:jids}
A key challenge in skill discovery is that high-frequency motion artifacts can be mistaken for behavioral diversity. During training, actions sampled from a stochastic policy can introduce high-frequency variations in the resulting motion. The skill discovery objective may reward these variations as diverse behaviors, allowing the policy to increase the diversity rewards through jittering rather than more meaningful behavioral variations. Similar tendencies to exploit high-frequency motion for diversity have also been observed in prior unsupervised skill discovery methods~\citep{atanassov2024constrained,cathomen2025d3}, while diversity objectives have also been shown to favor faster, short-period behaviors~\citep{park2026periodic}.

To mitigate this spurious diversity, we introduce Jitter-Induced Diversity Suppression (JIDS) to prevent high-frequency artifacts from being rewarded as behavioral diversity. JIDS applies a low-pass filter to the observations of the diffusion model and skill encoder, while the policy and discriminator observe the original unfiltered observations. The root trajectory is filtered in Euclidean space, while joint rotations are filtered on the rotation manifold using Slerp-based interpolation. In all experiments, JIDS uses a second-order Butterworth filter with a window size of $10$ timesteps and a cutoff frequency of $1\,\mathrm{Hz}$, applied to state sequences sampled at $30\,\mathrm{Hz}$. We will show that JIDS substantially reduces high-frequency motion artifacts and improves motion quality while largely preserving behavioral diversity.

\section{Downstream Control}
\label{sec:control}
The pretrained low-level policy provides a latent space of diverse motor skills. Using these skills for downstream control requires a mechanism to select skill latents that are appropriate for a given task. We consider two strategies for reusing skills: hierarchical control and zero-shot control.

\subsection{Hierarchical Control}
For hierarchical control, a task-specific high-level policy $\omega(\vc{z} | \vc{s}, \vc{g})$ is trained to select skill latents for the pretrained low-level policy $\pi(\vc{a} | \vc{s}, \vc{z})$ for each new task. The high-level policy receives as input the current state $\vc{s}$ and a task-specific goal $\vc{g}$. The low-level policy then executes actions conditioned on the skill latent selected by the high-level policy. Since skill latents lie on a unit hypersphere, the high-level output must lie on the same manifold. Following ASE~\citep{peng2022ASE}, we define the high-level action space as a Gaussian distribution $\bar{\vc{z}} \in \bar{\mathcal{Z}}$. A sampled action is then normalized and projected onto the skill latent space
$\vc{z} = \bar{\vc{z}} / \lVert \bar{\vc{z}} \rVert$,
and passed to the low-level policy. This parameterization retains a simple Gaussian action distribution while ensuring that each selected latent lies on the unit hypersphere.

\subsection{Zero-shot Control}
Hierarchical control requires training a task-specific high-level policy for each downstream task. Zero-shot control instead reuses the pretrained low-level policy without additional policy training. Since a fixed latent produces a consistent short-horizon behavior, it can be used as an open-loop skill command. A related skill-selection strategy was introduced by \citet{tirinzoni2025zero}, whereas DSD does not require forward-backward representation learning.

An offline trajectory pool is constructed by repeatedly rolling out the pretrained policy with fixed latents. Each entry is represented as
$\left(\tau^i,\vc{z}^i\right)$,
where
$\tau^i=\left\{\vc{s}_0^i,\vc{s}_1^i,\ldots,\vc{s}_T^i\right\}$
is generated using latent $\vc{z}^i$. The resulting dataset serves as a reusable pool of candidate behaviors. Downstream control can be performed by evaluating a task-specific fitness function $F(\tau^i, \vc{g})$ on each candidate trajectory $\tau^i$. The most suitable skill latent is determined by the trajectory that achieves the highest fitness:
\begin{equation}
\vc{z}^*
=
\vc{z}^{i^*},
\quad
i^*
=
\arg\max_i
F(\tau^i, \vc{g}).
\end{equation}
The selected latent $\vc{z}^*$ is then used to control the pretrained low-level policy for the desired task.    

\begin{figure}[t]
\centering
\includegraphics[width=0.9\linewidth]{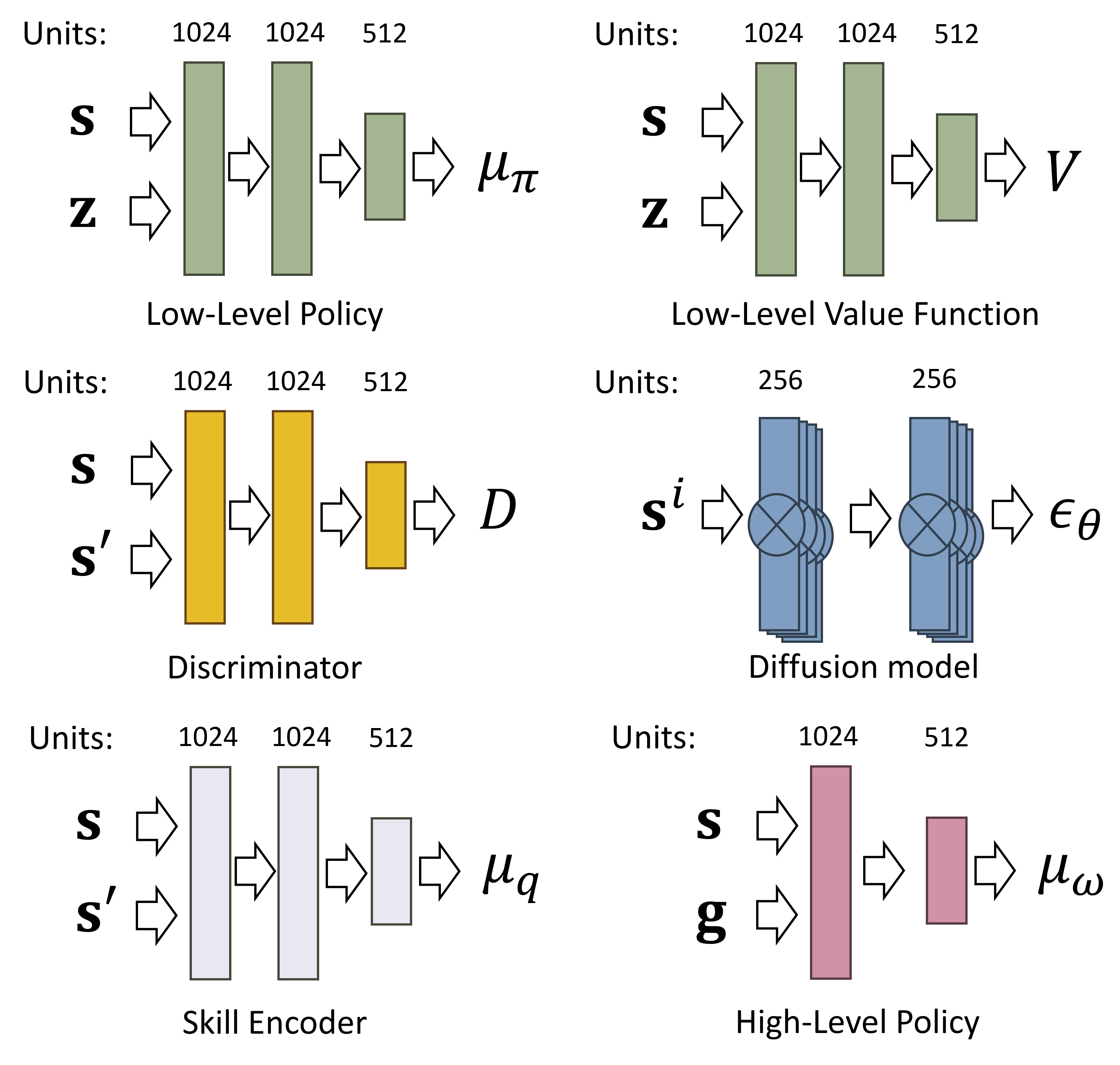} 
\caption{Network architectures used for our system. The low-level policy, low-level value function, discriminator, skill encoder and high-level policy are comprised of fully-connected layers with ReLU activations. The diffusion model consists of multi-head Transformer blocks with self-attention.}
\label{fig:policy}
\end{figure}

\begin{algorithm}[t]
\caption{Low-level Policy Training}
\label{alg:reusable_motor_skill_training}
\begin{algorithmic}[1]
\STATE{\textbf{Input:} motion dataset $\mathcal{M}$}
\STATE{$\epsilon_\theta \gets$ initialize diffusion model}
\STATE{$q \gets$ initialize skill encoder}
\STATE{$D \gets$ initialize discriminator}
\STATE{$\pi \gets$ initialize policy}
\STATE{$V \gets$ initialize value function}

\item[]
\WHILE{not done}
    \STATE{$\mathcal{B} \gets \emptyset$ initialize data buffer}
    \FOR{trajectory $j = 1,\ldots,m$}
        \STATE{sample latent $z \sim p(z)$}
        \STATE{collect trajectory $\tau^j = \{(s_t, a_t, s_{t+1})\}_{t=0}^{T-1}$ with $\pi$}
        \FOR{$t = 0,\ldots,T-1$}
            \STATE{$r_t^{\mathrm{ME}} \gets 0$ initialize marginal entropy reward}
            \FOR{diffusion steps $k \in \mathbb{K}$}
                \STATE{$\epsilon \sim \mathcal{N}(0, I)$}
                \STATE{$\epsilon_\theta\left( \vc{s}_t^k \right) \gets$ predict noise using Eq.~\ref{eq:epsilon_theta}}
                \STATE{$r_t^{\mathrm{ME}} \gets r_t^{\mathrm{ME}} + \left\lVert \epsilon_\theta\left( \vc{s}_t^k \right) - \epsilon \right\rVert^2$}
            \ENDFOR            
            \STATE{$r_t^{\mathrm{CE}} \gets \kappa\, \mu_q(\vc{s}_t)^T \vc{z}_t $} 
            \STATE{$r_t^{\mathrm{AMP}} \gets -\log\left(1 - D(\vc{s}_t, \vc{s}_{t+1})\right)$}
            \STATE{$r^\mathrm{DSD}_t \gets w^{\mathrm{ME}}r^{\mathrm{ME}}_t + w^{\mathrm{CE}} r^{\mathrm{CE}}_t$}
            \STATE{$r_t \gets w^{\mathrm{DSD}} r^{\mathrm{DSD}}_t + w^{\mathrm{AMP}} r_t^{\mathrm{AMP}}$}
            \STATE{record $(s_t, a_t, r_t, s_{t+1}, z)$ in $\tau^j$}
        \ENDFOR
        \STATE{store $\tau^j$ in $\mathcal{B}$}
        \ENDFOR
    \item[]
    \STATE{update $\epsilon_\theta$, $q$ and $D$ using data from $\mathcal{B}$}
    \STATE{update $V$ and $\pi$ using data from $\mathcal{B}$}
\ENDWHILE
\end{algorithmic}
\end{algorithm}

\section{Model Representation}
We apply DSD to develop reusable motor skills for 3D simulated humanoid characters. Our experiments are conducted on two characters: a 34-DOF humanoid without props, and a 37-DOF humanoid equipped with a sword and shield. The sword-and-shield character adds a 3-DOF joint in the right wrist to control a sword.

\subsection{States and Actions}
Following \citet{peng2022ASE}, the state $\vc{s}_t$ 
describes the configuration of the character's body:
\begin{itemize}
    \item Pelvis linear and angular velocities in the character's local coordinate frame.
    \item Pelvis height.
    \item Local joint rotations represented with the 6D rotation representation~\cite{zhou2020continuityrotationrepresentationsneural}.
    \item Local joint velocities.
    \item 3D positions of key joints in the character's local coordinate frame.
\end{itemize}
The action $\vc{a}_t$ specifies PD target joint rotations represented as 3D exponential maps~\cite{grassia1998practical}. 

\subsection{Network Architecture}
A schematic illustration of network architectures used in our system is provided as Fig.~\ref{fig:policy}.
The policy $\pi(\vc{a} | \vc{s}, \vc{z})$ is modeled with a three-layer MLP with linear output units. It takes the state $\vc{s}$ and skill latent $\vc{z}$ as input and outputs a Gaussian action distribution,
$\mathcal{N}\!\left(\mu_{\pi}(\vc{s}, \vc{z}), \Sigma_{\pi}\right)$.
The value function $V(\vc{s}, \vc{z})$ uses the same architecture and outputs a scalar value. The discriminator $D$ is represented by a two-layer MLP. The skill encoder uses the same architecture as the discriminator and outputs a unit-norm vector $\mu_q(\vc{s}_t)$.
All MLP-based networks use ReLU activations. Hidden layers use 1{,}024 units, except for the final hidden layer, which uses 512 units. The diffusion model is a two-layer Transformer with four attention heads with 256 hidden units that approximately yields 3M parameters. The number of diffusion timesteps is $M = 50$.

\subsection{Training Procedure}
The low-level policy is trained with proximal policy optimization (PPO)~\cite{schulman2017proximalpolicyoptimizationalgorithms}. Algorithm~\ref{alg:reusable_motor_skill_training} summarizes the training procedure. The diffusion model $\epsilon_\theta$ is trained on trajectory segments sampled from a separate replay buffer of recent policy rollouts, and is used to compute the marginal entropy reward $r_t^{\mathrm{ME}}$. At each timestep $t$, the policy receives the combined reward
$r_t = w^{\mathrm{DSD}} r_t^{\mathrm{DSD}} + w^{\mathrm{AMP}} r_t^{\mathrm{AMP}},$
where $r_t^{\mathrm{DSD}}$ promotes diverse and distinct skills, and $r_t^{\mathrm{AMP}}$ promotes natural behavior similar to those in the reference motion dataset $\mathcal{M}$. After collecting a batch of trajectories, PPO updates the policy using advantages estimated using GAE~\cite{schulman2015high}, and the value function is updated using TD($\lambda$)~\cite{sutton1998rl}. All networks are optimized with Adam~\cite{kingma2015adam}. The detailed parameters are provided in Appendix.~\ref{app:experimental setup}.

\section{Tasks}
We evaluate DSD on a suite of downstream tasks designed to cover different forms of skill reuse with both hierarchical and zero-shot control. In this section, we provide a high-level summary of each task. More detailed task descriptions are available in the Appendix~\ref{app:task_reward_designs}.

\subsection{Hierarchical Tasks}
For hierarchical control, tasks are designed to evaluate whether the learned low-level policy can support downstream behaviors that require compositions of different skills over extended horizons.

\paragraph{Steering.}
The steering task requires the character to move along a target direction at a fixed target speed. The character is rewarded for matching the target velocity and direction. 

\paragraph{Backflip}
In the backflip task, the character executes a backflip toward a specified direction. The reward encourages the character to move its root to a target height while moving along a target direction.

\paragraph{Location.}
The location task evaluates the model's ability to use locomotion skills to reach a target position. The character is required to a move to randomly placed target location.

\paragraph{Dodging.}
The dodging task requires the character to avoid opponents that are running towards it. The character is rewarded for successfully dodging the opponent and penalized for collisions.

\paragraph{Shield Bash.}
In the shield bash task, the character must knock down incoming opponents. The reward encourages fast shield strikes on the opponent.

\paragraph{Beat Saber.}
The Beat Saber task evaluates whether the model can produce sword-swing motions with precise timing and spatial accuracy. The objective is for the character to slash incoming blocks with a sword.

\begin{figure}[t]
    \centering
    \begin{minipage}{\linewidth}
        \centering

        \begin{subfigure}{0.95\linewidth}
            \centering
            \includegraphics[width=0.19\linewidth]{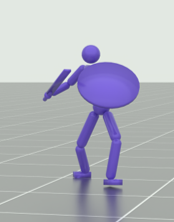}\hfill
            \includegraphics[width=0.19\linewidth]{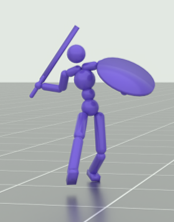}\hfill
            \includegraphics[width=0.19\linewidth]{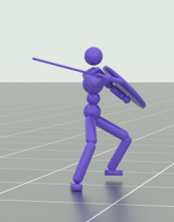}\hfill
            \includegraphics[width=0.19\linewidth]{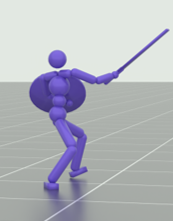}\hfill
            \includegraphics[width=0.19\linewidth]{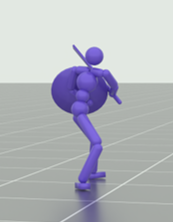}
            \caption{Sword Attack}
        \end{subfigure}
        \vspace{0.2em}
        \begin{subfigure}{0.95\linewidth}
            \centering
            \includegraphics[width=0.19\linewidth]{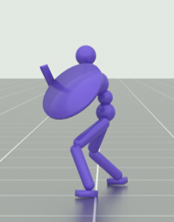}\hfill
            \includegraphics[width=0.19\linewidth]{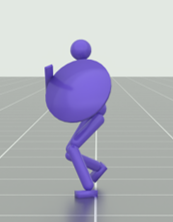}\hfill
            \includegraphics[width=0.19\linewidth]{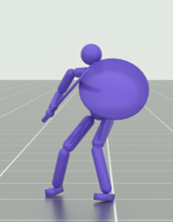}\hfill
            \includegraphics[width=0.19\linewidth]{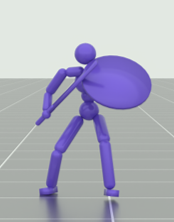}\hfill
            \includegraphics[width=0.19\linewidth]{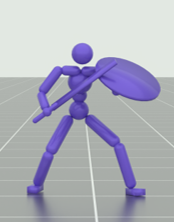}
            \caption{Turn}
        \end{subfigure}

        \vspace{0.2em}

        \begin{subfigure}{0.95\linewidth}
            \centering
            \includegraphics[width=0.19\linewidth]{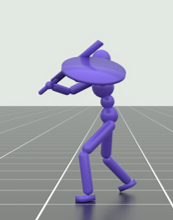}\hfill
            \includegraphics[width=0.19\linewidth]{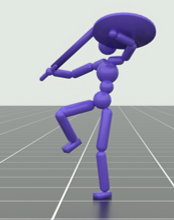}\hfill
            \includegraphics[width=0.19\linewidth]{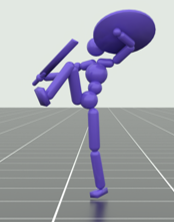}\hfill
            \includegraphics[width=0.19\linewidth]{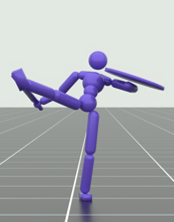}\hfill
            \includegraphics[width=0.19\linewidth]{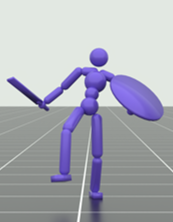}
            \caption{Kick}
        \end{subfigure}
        \vspace{0.2em}        \begin{subfigure}{0.95\linewidth}
            \centering
            \includegraphics[width=0.19\linewidth]{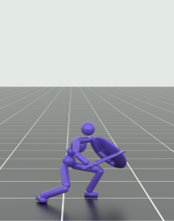}\hfill
            \includegraphics[width=0.19\linewidth]{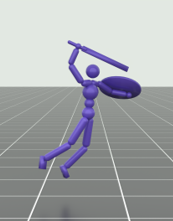}\hfill
            \includegraphics[width=0.19\linewidth]{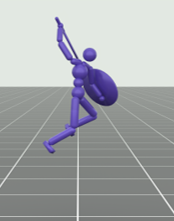}\hfill
            \includegraphics[width=0.19\linewidth]{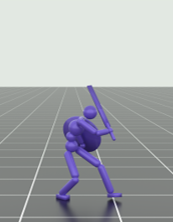}\hfill
            \includegraphics[width=0.19\linewidth]{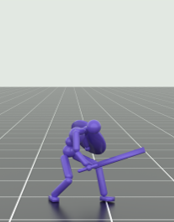}
            \caption{Jump and Slash}
        \end{subfigure}

        \vspace{0.2em}

        \begin{subfigure}{0.95\linewidth}
            \centering
            \includegraphics[width=0.19\linewidth]{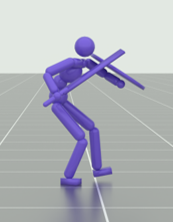}\hfill
            \includegraphics[width=0.19\linewidth]{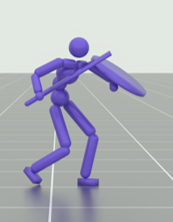}\hfill
            \includegraphics[width=0.19\linewidth]{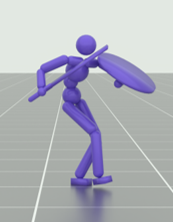}\hfill
            \includegraphics[width=0.19\linewidth]{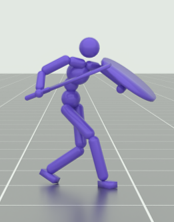}\hfill
            \includegraphics[width=0.19\linewidth]{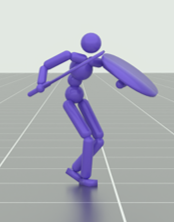}
            \caption{Strafe}
        \end{subfigure}
    \end{minipage}

    \caption{Physically simulated character performing skills produced by random latents. The pretrained low-level policy generates a diverse set of reusable behaviors within a single latent-conditioned model.}
    \label{fig:learned_skills}
\end{figure}

\begin{figure}[t]
    \centering
    \begin{minipage}{0.95\linewidth}
        \centering
        \begin{subfigure}{0.95\linewidth}
            \centering
            \includegraphics[width=\linewidth]{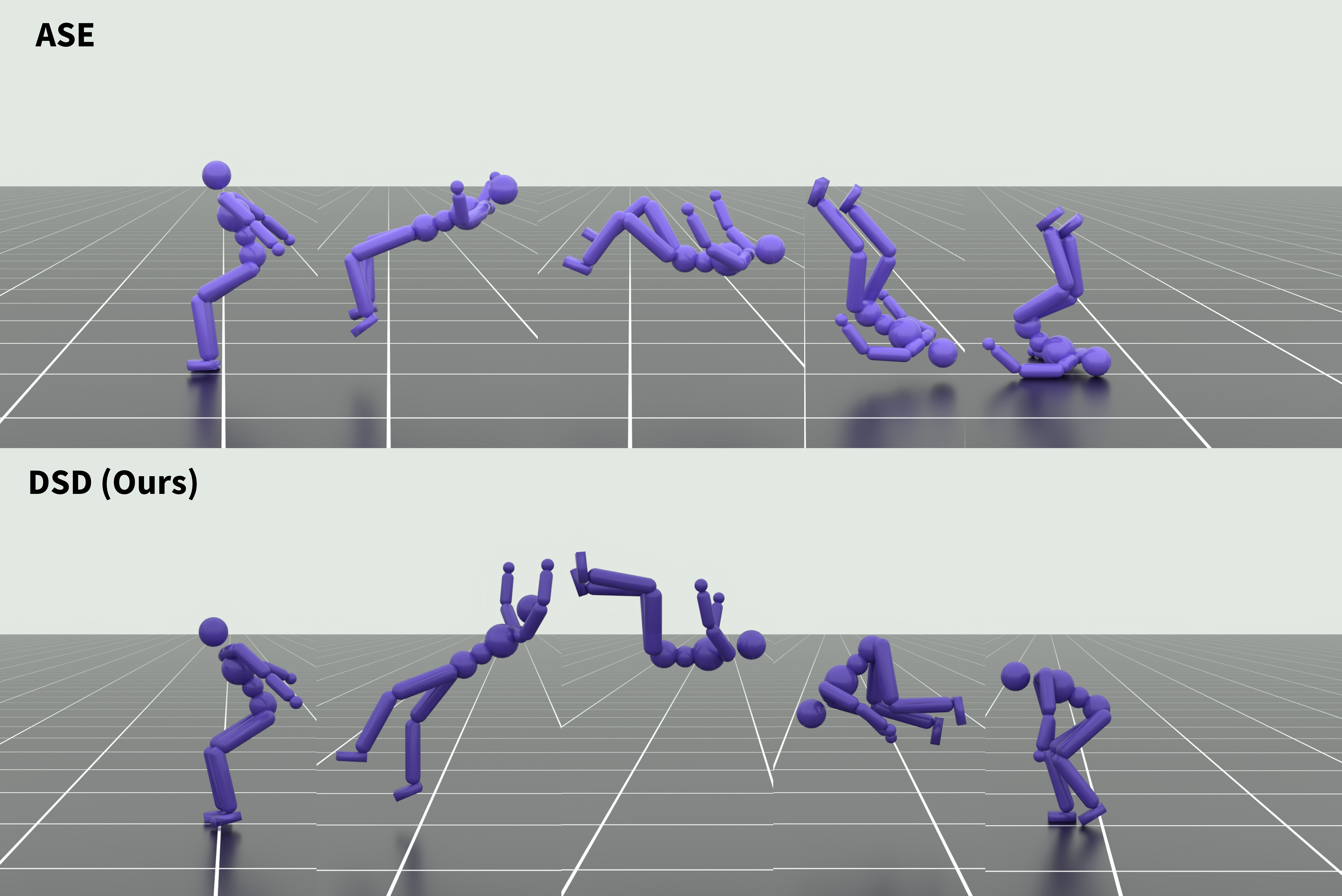}
            \caption{Backflip}
        \end{subfigure}
        \vspace{0.2em}
        \begin{subfigure}{0.95\linewidth}
            \centering
            \includegraphics[width=\linewidth]{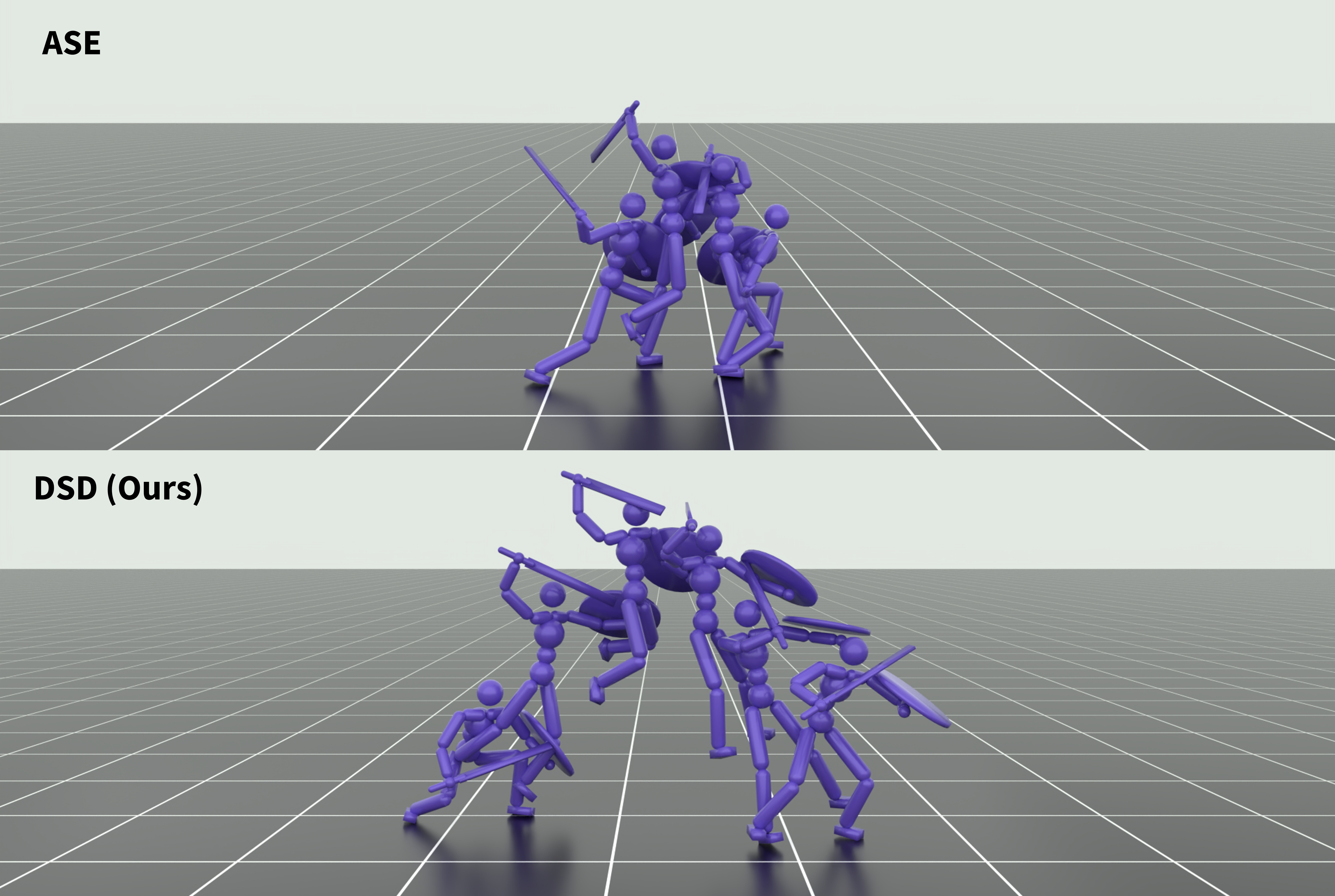}
            \caption{Jump}
        \end{subfigure}
    \end{minipage}
    \caption{Visual comparison of learned motion skills. DSD produces challenging dynamic skills, including backflips and jumps, whereas ASE fails to complete these motions.}
    \label{fig:vis_comparison}
\end{figure}

\subsection{Zero-shot Tasks}
The zero-shot tasks evaluate whether the pretrained skills can be used to perform new tasks without additional training or task-specific data. A fixed trajectory dataset is recorded beforehand by rolling out the pretrained policy with randomly sampled latents. For each task, the fitness function is evaluated only on this pre-recorded dataset, and the latent paired with the highest-fitness trajectory is selected. In our experiments, we consider two types of fitness functions, Sum and Max, which are used depending on the characteristics of a given task.

The Sum fitness function is used for tasks where performance should be assessed over the full trajectory. For a candidate trajectory $\tau^i$, the fitness is computed as the sum of per-timestep task rewards:
\begin{equation}
F^{\mathrm{sum}}(\tau^i, \vc{g})
=
\sum_{t=0}^{T}
r_t^{G}(\vc{s}_t^i, \vc{g}).
\end{equation}
Here, $r_t^{G}$ denotes the task reward at timestep $t$ for a fixed goal $\vc{g}$. This form is used for tasks where the selected latent should maintain progress toward the goal throughout the horizon, such as strafe, location, and combat tasks.

For tasks where success is determined by the best state reached within the trajectory, we use a Max fitness function. For a candidate trajectory $\tau^i$, the fitness is computed as the maximum task reward over the horizon:
\begin{equation}
F^{\mathrm{max}}(\tau^i, \vc{g})
=
\max_{0 \leq t \leq T}
r_t^{G}(\vc{s}_t^i, \vc{g}).
\end{equation}
This form is used for running, reaching, jumping and ducking tasks, where the selected trajectory only needs to reach the desired state at some point within the horizon.

\begin{figure*}[t]
\centering
\includegraphics[width=0.95\linewidth]{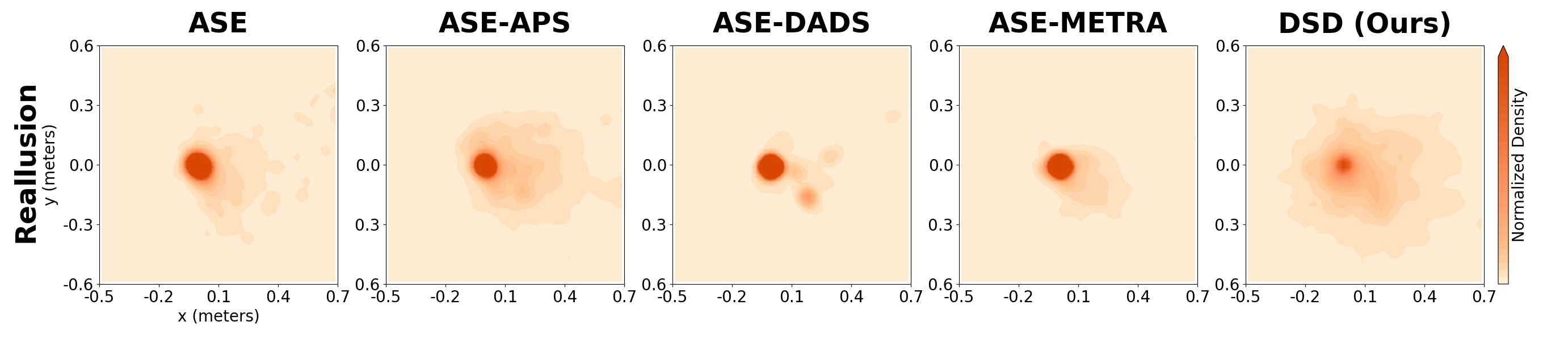} 
\includegraphics[width=0.95\linewidth]{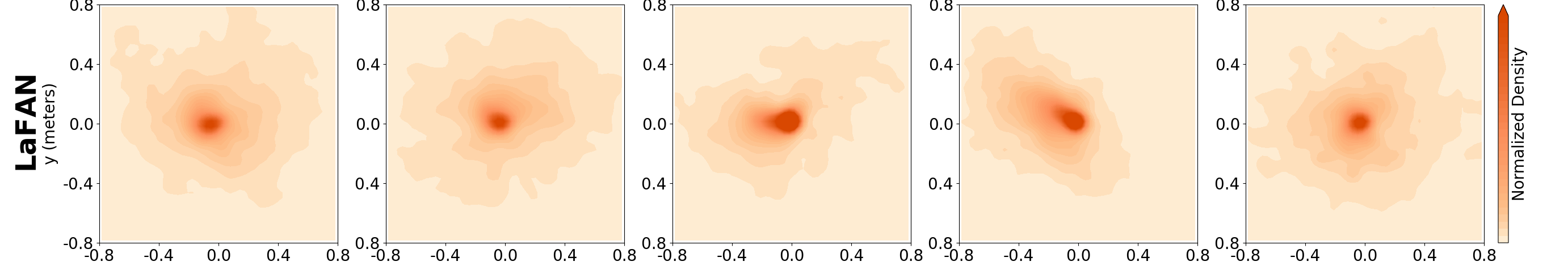} 
\includegraphics[width=0.95\linewidth]{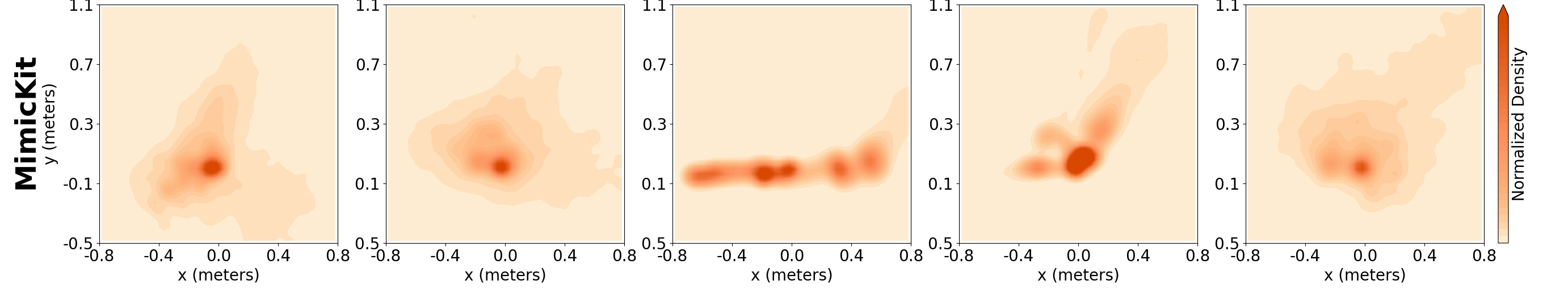}

\caption{2D root-position heatmaps for different skill discovery methods applied to the Reallusion, LaFAN, and MimicKit datasets.
Each heatmap is computed from 1{,}024 trajectories of 10 seconds in length.
Across datasets, DSD generally exhibits a broader spatial coverage, whereas the baseline methods tend to be more concentrated in certain regions, often near the initial location.}
\label{fig:root_heatmap}
\end{figure*}

\section{Experiments}
Our experiments are designed to evaluate DSD's ability to learn a broad repertoire of human-like motor skills, and the effectiveness of the learned skills when applied to diverse downstream tasks. We evaluate DSD's ability to automatically discover diverse behaviors by training models on 3 distinct motion datasets of varying scales and diversity. The reusability of the learned skills is then evaluated on downstream tasks through hierarchical control and zero-shot control. Through our ablation experiments, we isolate the contributions of key design decisions. The motions produced by DSD are best viewed in the supplementary video.

\paragraph{Baselines.}
We compare DSD with prior skill-discovery methods under the same training setting. ASE serves as the primary baseline since ASE and DSD use a low-level policy conditioned on the character state and a skill latent, and both incorporate an adversarial motion prior~\citep{peng2022ASE}. Their difference lies in the skill discovery objective: ASE uses a DIAYN-based objective, whereas DSD uses the proposed diffusion-based objective~\citep{eysenbach2018diversity}. This comparison isolates the effect of replacing DIAYN with the proposed DSD objective. To further compare different skill discovery objectives within the same framework, we replace the DIAYN objective in ASE with alternative skill discovery objectives. APS~\citep{liu2021aps} and DADS~\citep{sharma2019dynamics} are selected as tractable approaches to state-entropy estimation, using particle-based estimation and skill-conditioned dynamics, respectively. METRA~\citep{park2024metra} provides a complementary comparison through a metric-aware representation that preserves temporal distances between states. The latent-conditioned policy and adversarial motion prior remain unchanged across these variants, which are denoted as ASE-APS, ASE-DADS, and ASE-METRA. This design focuses the comparison on alternative skill discovery objectives that can be evaluated within the same low-level policy framework.

\paragraph{Motion Datasets.}
The experiments use three motion datasets that differ substantially in scale and motion characteristics. Reallusion serves as the main evaluation dataset, containing 187 motion clips totaling approximately 30 minutes, with gladiator-style motions that are performed with a sword and shield~\citep{reallusion2022}. This dataset supports weapon-based combat skills and provides the most direct comparison to ASE, which was evaluated on the same dataset \citep{peng2022ASE}. LaFAN1 evaluates the scalability of DSD  to learn from diverse everyday and expressive motions, with approximately 160 minutes of motion data including locomotion, gestures, and dancing~\citep{harvey20robust}.
MimicKit tests skill discovery under limited data, containing 27 short clips that total 2.5 minutes of motion and including highly dynamic behaviors such as backflips and jump kicks~\citep{MimicKitPeng2025}.

\begin{table*}[t]
\centering
\setlength{\tabcolsep}{5pt}
\caption{Comparison of motion quality and diversity with different skill discovery methods across datasets. Five models initialized with different random seeds are trained for each method. Lower FID indicates better agreement with the reference motion distribution. Higher root motion diversity (Div-R) and local joint motion diversity (Div-J) indicate broader behavioral diversity. Best results are highlighted in gray and second-best are underlined.}
\label{tab:fid_div_results}
\begin{tabular}{lccccccccc}
\toprule
& \multicolumn{3}{c}{\textbf{Reallusion}}
& \multicolumn{3}{c}{\textbf{LaFAN1}}
& \multicolumn{3}{c}{\textbf{MimicKit}} \\
\cmidrule(lr){2-4} \cmidrule(lr){5-7} \cmidrule(lr){8-10}

\textbf{Method}
& \textbf{FID} $\downarrow$
& \textbf{Div-R} $\uparrow$
& \textbf{Div-J} $\uparrow$
& \textbf{FID} $\downarrow$
& \textbf{Div-R} $\uparrow$
& \textbf{Div-J} $\uparrow$
& \textbf{FID} $\downarrow$
& \textbf{Div-R} $\uparrow$
& \textbf{Div-J} $\uparrow$ \\

\midrule

\textbf{ASE}
& $2.47^{\pm 0.37}$
& $0.113^{\pm 0.010}$
& $0.144^{\pm 0.002}$
& $3.08^{\pm 1.21}$
& $0.167^{\pm 0.002}$
& $0.197^{\pm 0.003}$
& $0.61^{\pm 0.23}$
& $0.143^{\pm 0.003}$
& $0.311^{\pm 0.039}$ \\

\textbf{ASE-APS}
& \underline{$2.02^{\pm 0.21}$}
& \underline{$0.123^{\pm 0.005}$}
&  \underline{$0.169^{\pm 0.003}$}
& \underline{$2.28^{\pm 0.15}$}
& \underline{$0.173^{\pm 0.004}$}
& $0.202^{\pm 0.002}$
& \underline{$0.31^{\pm 0.26}$}
& \underline{$0.174^{\pm 0.009}$}
& \underline{$0.352^{\pm 0.032}$} \\

\textbf{ASE-DADS}
& $16.73^{\pm 5.55}$
& $0.040^{\pm 0.033}$
& \cellcolor{gray!20}{$0.193^{\pm 0.122}$}
& $64.34^{\pm 34.76}$
& $0.075^{\pm 0.086}$
& \cellcolor{gray!20}{$0.227^{\pm 0.043}$}
& $4.62^{\pm 7.22}$
& $0.043^{\pm 0.057}$
& $0.256^{\pm 0.092}$ \\

\textbf{ASE-METRA}
& $12.39^{\pm 7.75}$
& $0.068^{\pm 0.062}$
& $0.083^{\pm 0.050}$
& $4.16^{\pm 1.66}$
& $0.150^{\pm 0.031}$
& $0.179^{\pm 0.028}$
& $1.54^{\pm 1.65}$
& $0.137^{\pm 0.025}$
& $0.287^{\pm 0.114}$ \\

\midrule

\textbf{DSD (Ours)}
& \cellcolor{gray!20}{$1.64^{\pm 0.10}$}
& \cellcolor{gray!20}{$0.130^{\pm 0.003}$}
& \underline{$0.169^{\pm 0.004}$}
& \cellcolor{gray!20}{$1.88^{\pm 0.04}$}
& \cellcolor{gray!20}{$0.178^{\pm 0.003}$}
& \underline{$0.204^{\pm 0.001}$}
& \cellcolor{gray!20}{$0.29^{\pm 0.25}$}
& \cellcolor{gray!20}{$0.186^{\pm 0.010}$}
& \cellcolor{gray!20}{$0.366^{\pm 0.024}$} \\

\bottomrule
\end{tabular}
\end{table*}

\begin{figure*}[t]
\centering
\includegraphics[width=0.95\linewidth]{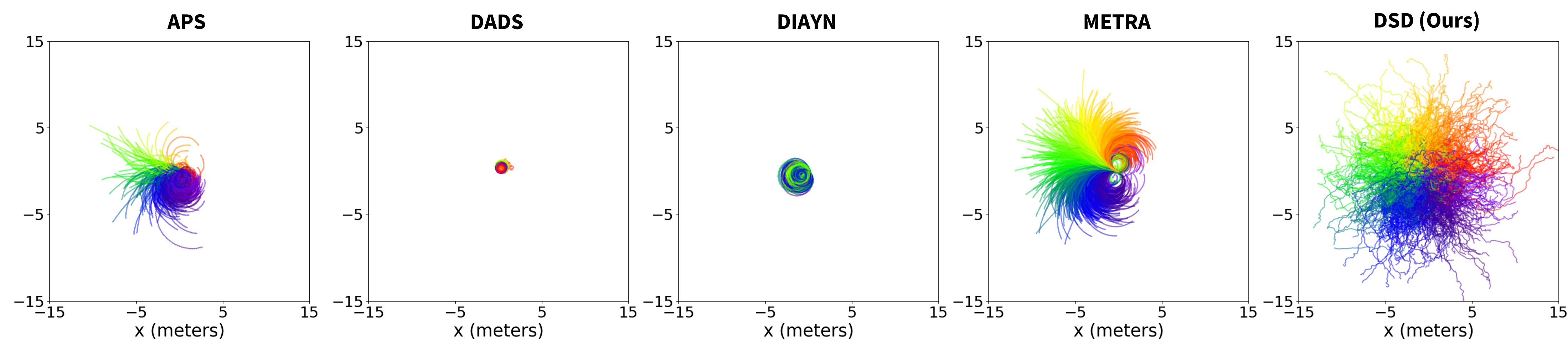} 
\caption{2D root trajectories produced by pure unsupervised learning policies trained with various methods. Each method is visualized using 1,024 trajectories of 10 seconds each.
DSD exhibits broader and more diverse behaviors compared to the baselines.
}
\label{fig:pure_2d}
\end{figure*}

\subsection{Learned Skills}
Effective skill discovery should produce a diverse repertoire of distinct behaviors. Figure~\ref{fig:learned_skills} shows representative examples of skills from a DSD model trained on the Reallusion dataset. Randomly sampled latents produce a variety of locomotion, jumps, kicks, and combat maneuvers, illustrating the range of behaviors represented by the DSD model.

To evaluate the effectiveness of various skill discovery methods, we analyze whether a method can automatically discover challenging dynamic behaviors. Motions that involve significant flight phases, such as backflips and large jumps, require precise timing and coordinated full-body movement, where small errors can prevent successful completion. Prior work has introduced specialized mechanisms, such as nonphysical residual forces,
to facilitate the acquisition of agile motions~\citep{dou2022case}. In our experiments, we compare the effectiveness of DSD and ASE in discovering these challenging dynamic skills without additional assistive forces. Experiments are conducted using the MimicKit dataset and the Reallusion. Figure~\ref{fig:vis_comparison} shows representative dynamic motions discovered by DSD and ASE. On MimicKit, DSD produces complete backflips, whereas ASE initiates the rotation but fails to complete the full flip. On Reallusion, both methods produce jumping behaviors, while DSD exhibits substantially larger jumps. These examples show that DSD can effectively discover challenging dynamic behaviors without additional mechanisms specialized for agile motions.

\begin{figure*}[t]
\begin{minipage}{0.95\linewidth}

\centering
    \vspace{0.5em}
    \begin{subfigure}{0.48\linewidth}
        \centering
        \includegraphics[width=\linewidth, trim=0 0 0 6mm, clip
        ]{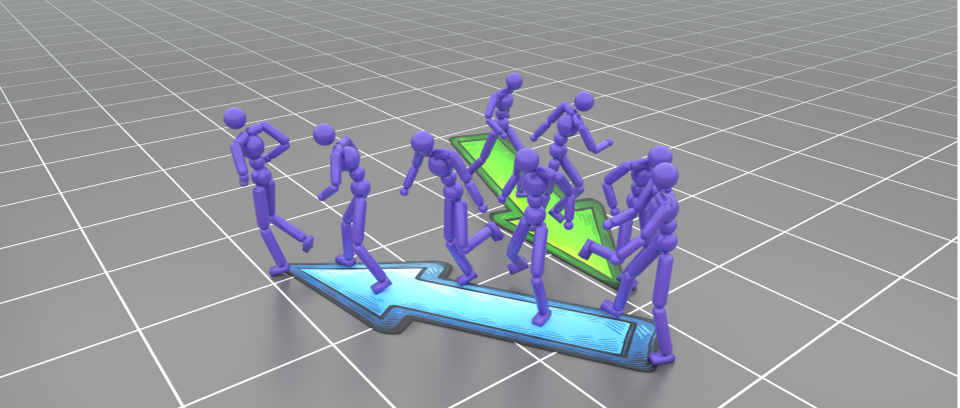}
        \caption{Steer}
    \end{subfigure}
    \hfill
    \begin{subfigure}{0.48\linewidth}
        \centering
        \includegraphics[width=\linewidth, trim=0 0 0 6mm, clip]{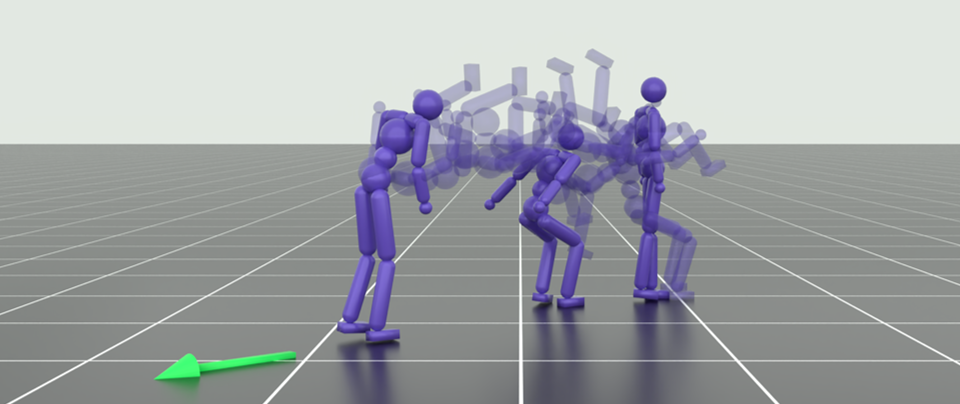}
        \caption{Backflip}
    \end{subfigure}  
    
    \centering
    \vspace{0.5em}
    \begin{subfigure}{0.48\linewidth}
        \centering
        \includegraphics[width=\linewidth, trim=0 0 0 4mm, clip]{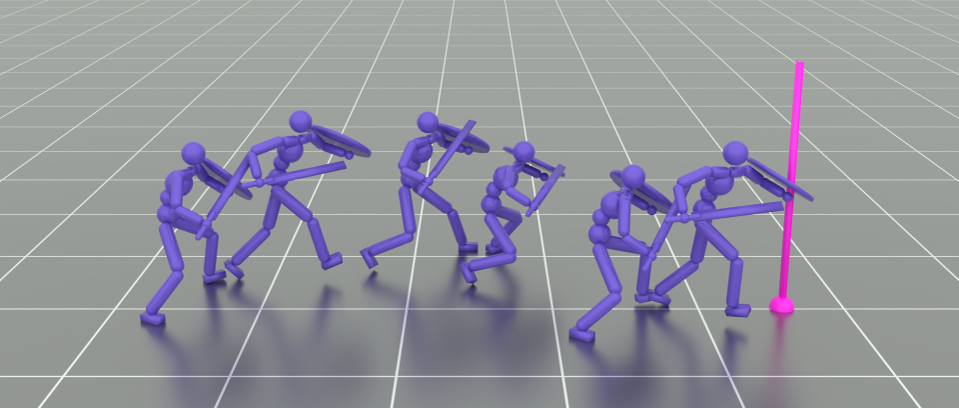}
        \caption{Location}
    \end{subfigure}
    \hfill
    \begin{subfigure}{0.48\linewidth}
        \centering
        \includegraphics[width=\linewidth, trim=0 0 0 4mm, clip]{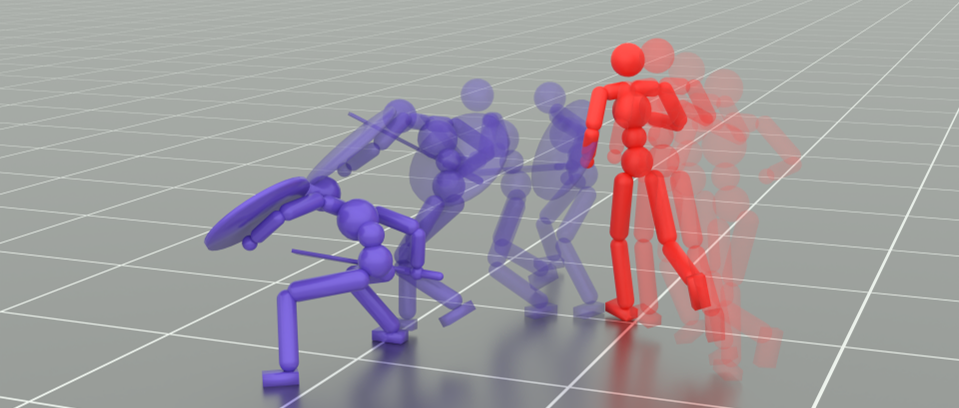}
        \caption{Dodging}
    \end{subfigure}  
    
     \vspace{0.5em}
    \begin{subfigure}{0.48\linewidth}
        \centering
        \includegraphics[width=\linewidth, trim=0 0 0 4mm, clip]{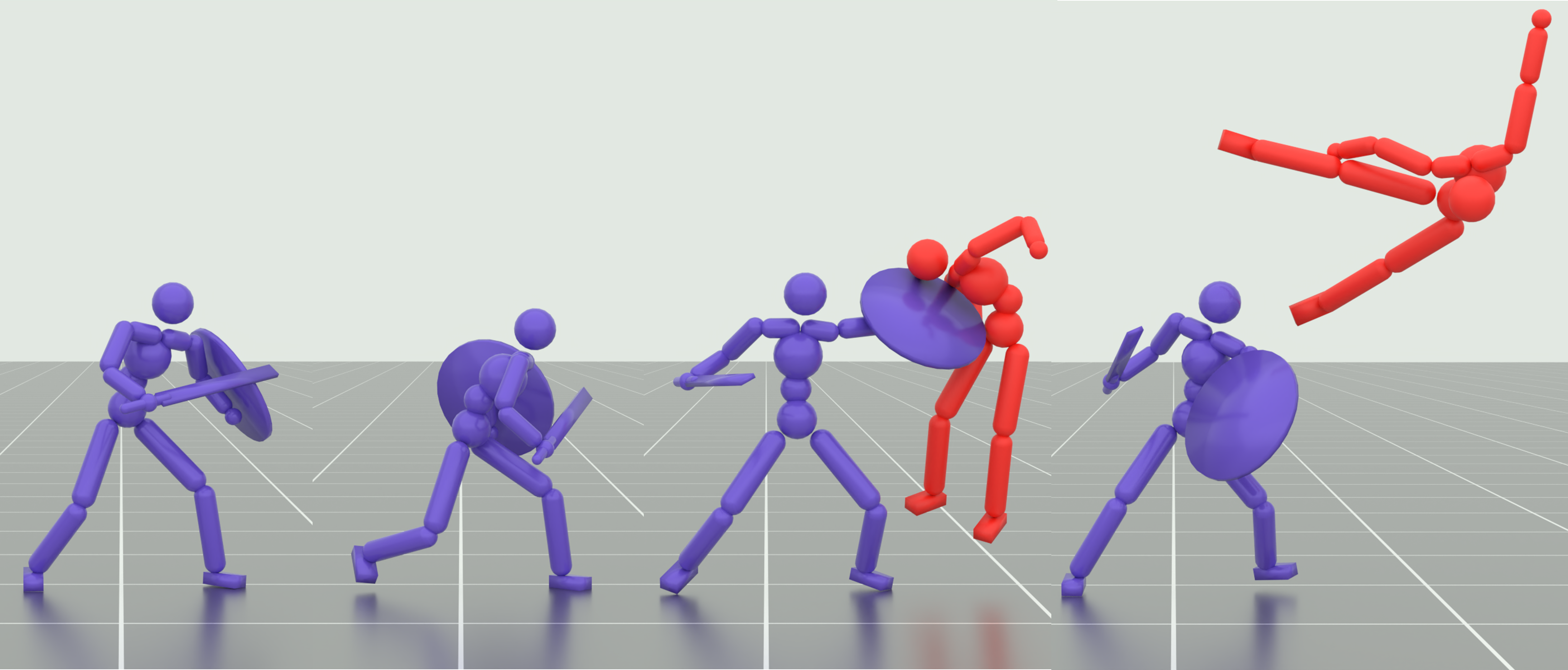}
        \caption{Shield Bash}
    \end{subfigure}
    \hfill
    \begin{subfigure}{0.48\linewidth}
    \centering
    \includegraphics[width=\linewidth, trim=0 0 0 4mm, clip]{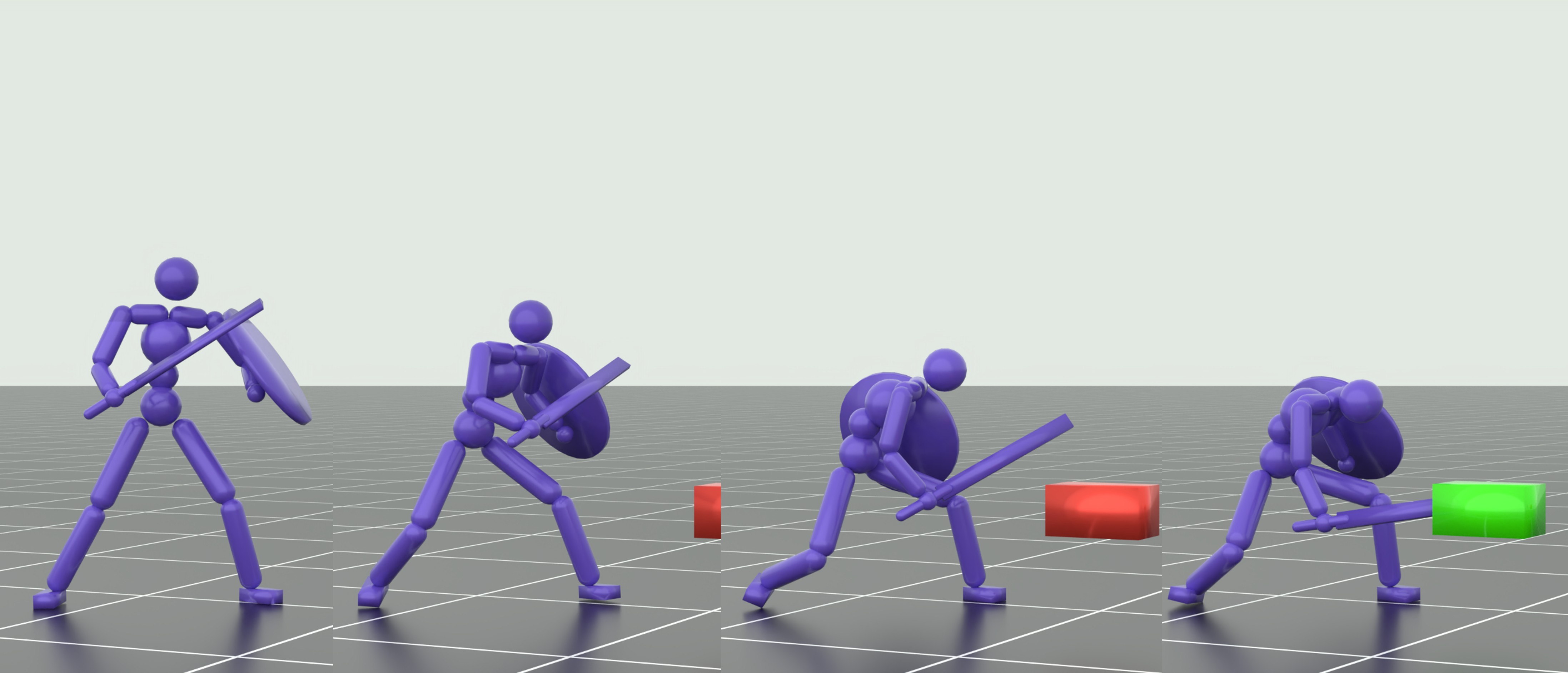}
    \caption{BeatSaber}
    \end{subfigure}
        
    \end{minipage}
    \caption{Simulated character performing downstream tasks using hierarchical control. The pretrained low-level policy provides reusable motor skills that can be effectively used for different downstream tasks, including locomotion and combat maneuvers.}
    \label{fig:tasks}
\end{figure*}

\subsection{Behavioral Diversity}
\label{subsec:result-div}
To evaluate behavioral diversity, we consider the root trajectories and the local body motions produced by a control model. Random trajectories are generated by conditioning the low-level policy on randomly sampled skill latents. Motion statistics are then calculated from these trajectories and compared with a number of prior skill discovery methods.

\paragraph{Root Motion Coverage.}
Root motions provide a useful quantity for analyzing motion diversity, since the movement of the root is a consequence of the full-body movements of a character.
Figure~\ref{fig:root_heatmap} visualizes the 2D distribution of root positions from 1{,}024 randomly sampled trajectories for each method, where each trajectory is 10 seconds long. The baseline methods generate trajectories that concentrate mostly near the initial starting location. In contrast, DSD generates trajectories that cover a wider range of directions and distances. This pattern indicates that skills learned through DSD can produce more diverse trajectories.

\paragraph{Root and Joint Diversity.}
Diversity across generated motions is evaluated using root and joint diversity metrics. Before computing the diversity metrics, each 30-frame motion segment is canonicalized with respect to its first frame by translating the initial root position to the origin and aligning the initial heading direction to a fixed forward direction. Root diversity then measures the pairwise differences between the resulting horizontal root trajectories, while joint diversity measures pairwise differences between the local body motions. 1{,}024 trajectories are generated per method, with each trajectory spanning 25 seconds. The skill latent is resampled every 5--6 seconds. Diversity is computed from randomly selected pairs of 30-frame windows. As shown in Table~\ref{tab:fid_div_results}, DSD exhibits the highest root-motion diversity on all three datasets. It also obtains the highest joint-motion diversity on MimicKit and the second-highest values on Reallusion and LaFAN1. These results show that DSD expands spatial coverage while maintaining substantial variations in local body motions.

\paragraph{Pure Unsupervised RL}
To isolate the effect of the diffusion-based skill discovery objective, we train models without the adversarial motion-prior reward $r_t^{\mathrm{AMP}}$, using only the skill discovery objective of each method. Early termination remains unchanged from the full framework and is applied when any body part other than the feet contacts the ground. Baseline methods often remain close to the initial location and produce repetitive movement patterns. In contrast, DSD produces substantially broader spatial coverage, as illustrated by the root trajectories in Figure~\ref{fig:pure_2d}. The supplementary video additionally shows more varied body motions.

\subsection{Motion FID}
For downstream applications, diversity alone may not be sufficient; the learned skills should also produce naturalistic behaviors. We therefore evaluate generated motion quality using Fr\'{e}chet Inception Distance (FID). FID compares the feature distributions of the generated motions to a reference dataset, where lower values indicate smaller distributional differences from the dataset. In our experiments, each motion trajectory is generated by conditioning the low-level policy on a random skill latent. FID is computed using 3{,}000 motion subsequences sampled from the resulting trajectories. Motion features are extracted using a transformer-based ACTOR-style autoencoder~\cite{petrovich21Actor}.

As shown in Table~\ref{tab:fid_div_results}, DSD achieves the lowest FID across all three datasets. These results indicate that DSD achieves broader behavioral diversity while remaining closely aligned with the reference motion distribution. The comparison with ASE-DADS further highlights the importance of considering motion quality together with diversity. On Reallusion and LaFAN1, ASE-DADS achieves the highest joint motion diversity, but this is accompanied by substantially higher FID values of $16.73$ and $64.34$, compared with $1.64$ and $1.88$ for DSD. This high joint-motion diversity is accompanied by substantially poorer agreement with the reference distribution. In contrast, DSD achieves the lowest FID and highest root-motion diversity across all three datasets, while obtaining either the highest or second-highest joint-motion diversity.

\begin{figure}[t]
\centering
\includegraphics[width=0.95 \columnwidth]{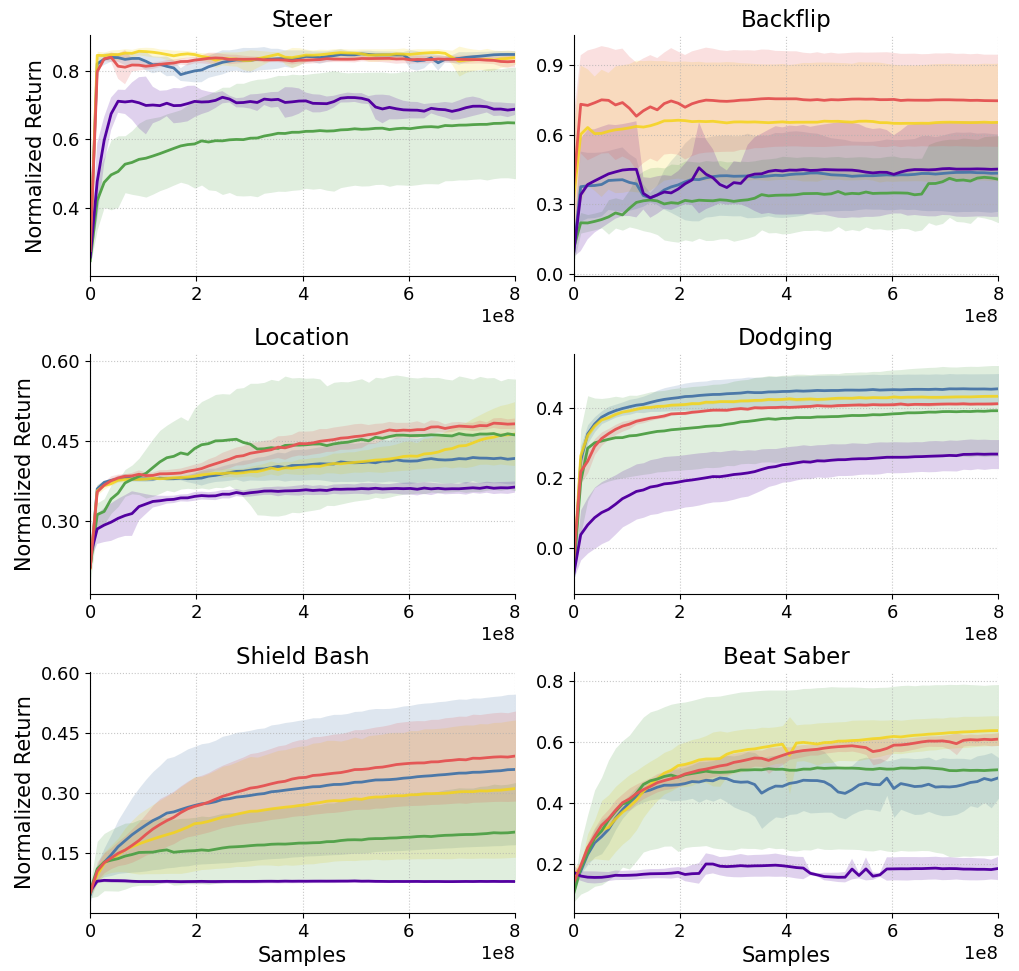}
\begin{subfigure}{0.95\linewidth}
    \centering
    \includegraphics[width=\linewidth]{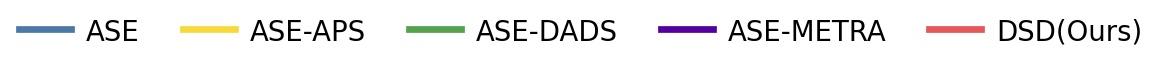}
\end{subfigure}    
\caption{
Learning curves for downstream tasks in the hierarchical control setting. Each method uses a pretrained low-level policy, while a task-specific high-level policy is trained to select skill latents.
DSD exhibits strong performance across tasks that require goal-directed movements, dynamic body motions, and object interactions.
}
\label{fig:hlc_learning_curve}
\end{figure}

\begin{figure*}[t]
    \centering
    \begin{minipage}{0.95\linewidth}
        \centering

        \begin{subfigure}{0.48\linewidth}
            \centering
            \includegraphics[
                width=\linewidth,
                trim=0 0 0 8mm,
                clip
            ]{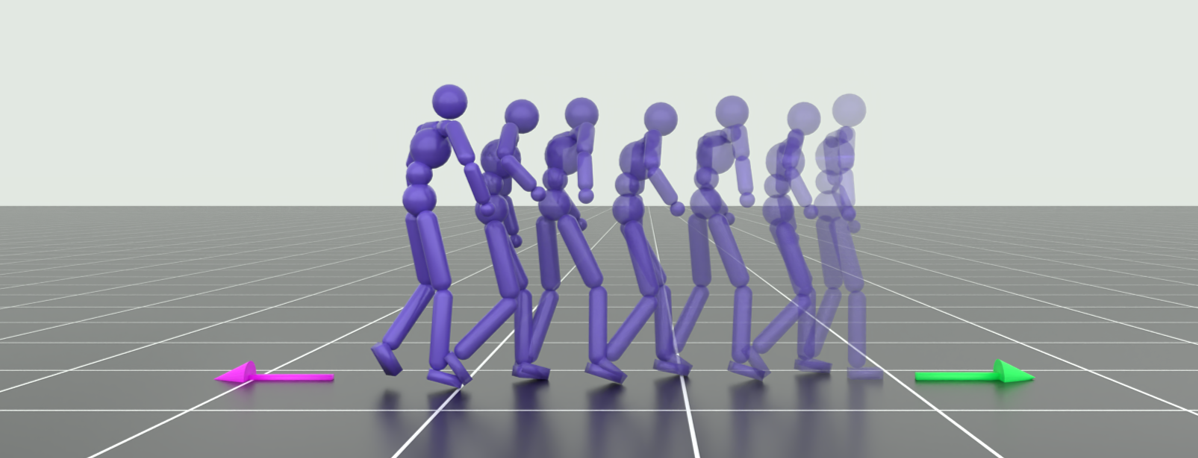}
            \caption{Strafe}
        \end{subfigure}
        \hfill
        \begin{subfigure}{0.48\linewidth}
            \centering
            \includegraphics[
                width=\linewidth,
                trim=0 0 0 8mm,
                clip
            ]{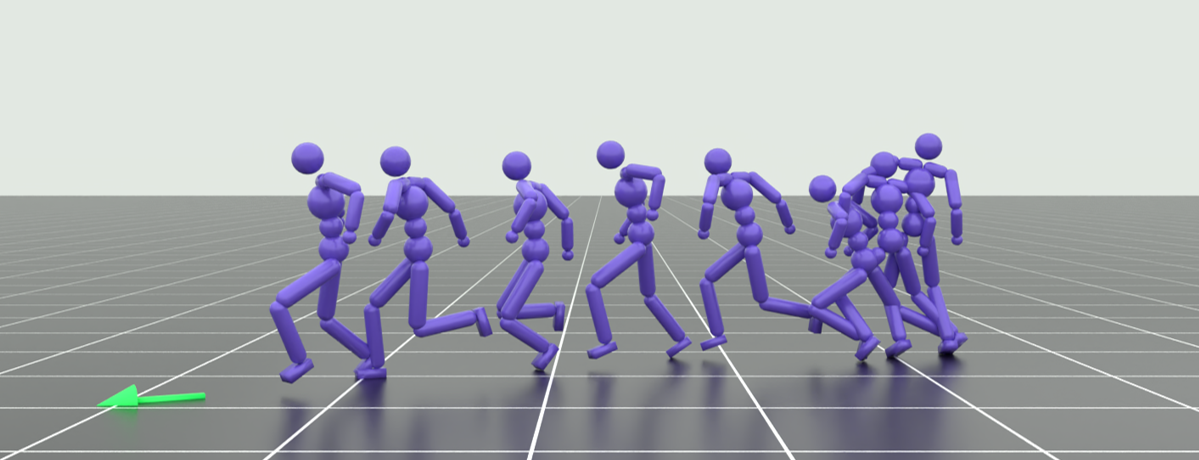}
            \caption{Run}
        \end{subfigure}

        \vspace{0.5em}

        \begin{subfigure}{0.48\linewidth}
            \centering
            \includegraphics[
                width=\linewidth,
                trim=0 0 0 12mm,
                clip
            ]{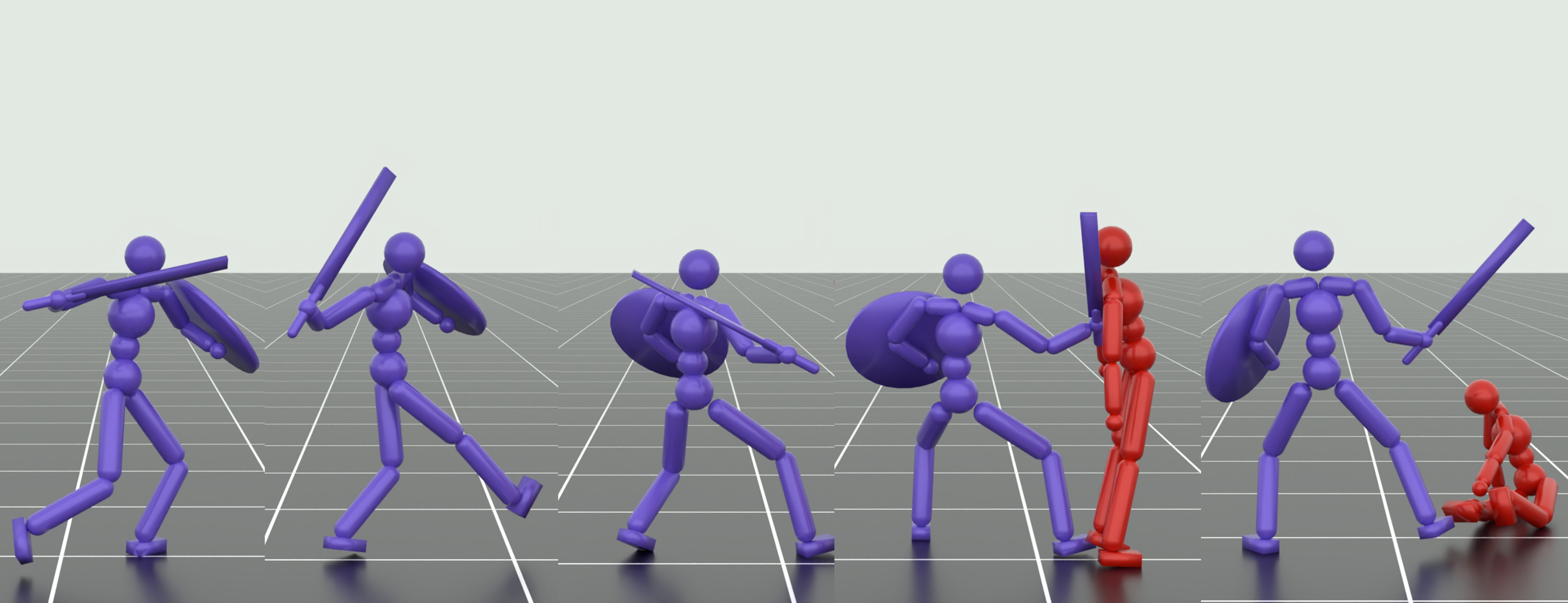}
            \caption{Sword Attack}
        \end{subfigure}
        \hfill
        \begin{subfigure}{0.48\linewidth}
            \centering
            \includegraphics[
                width=\linewidth,
                trim=0 0 0 12mm,
                clip
            ]{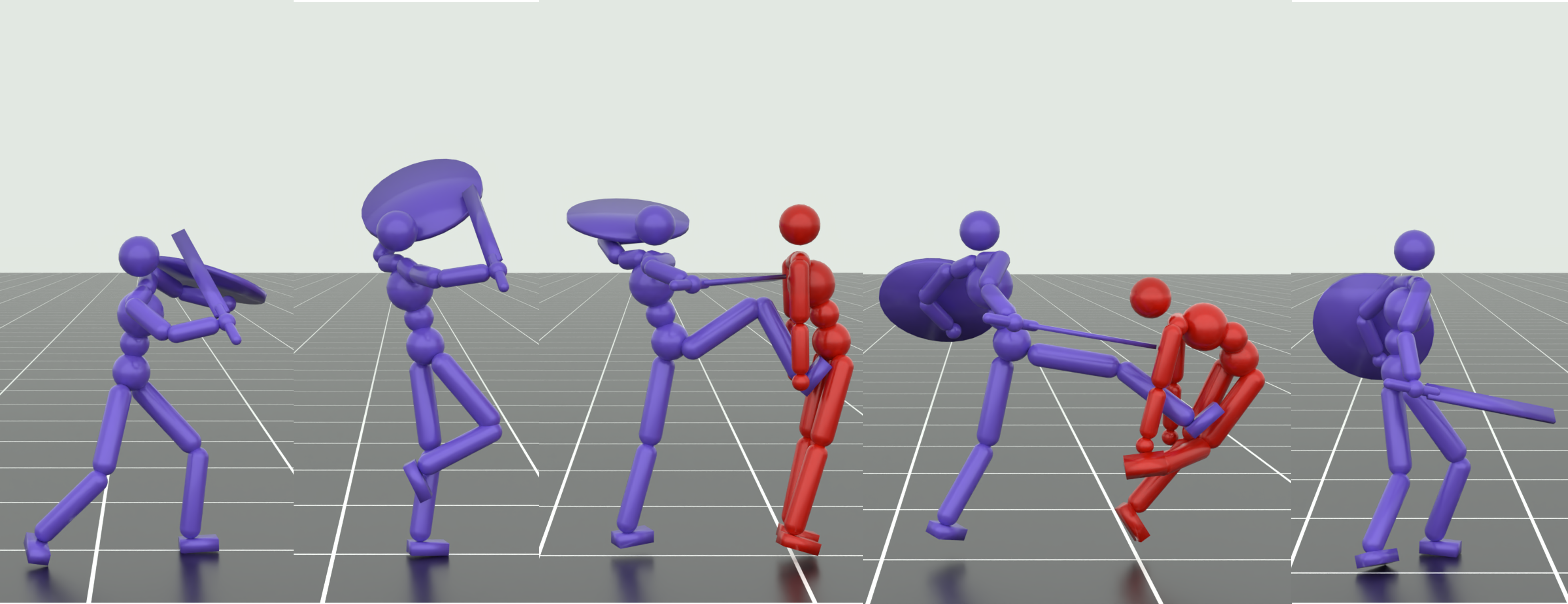}
            \caption{Kick Attack}
        \end{subfigure}
    \end{minipage}

    \caption{Simulated character performing downstream tasks using zero-shot control. Offline trajectories collected from the pretrained low-level policy form a candidate pool, and task-specific fitness functions are used to select skills for downstream tasks, ranging from locomotion to combat.}
    \label{fig:zeroshot_tasks}
\end{figure*}

\subsection{Hierarchical Control Performance}
Hierarchical control evaluates whether the learned skills provide behaviors that can be reused by task-specific high-level policies. We design six tasks that require substantially different behaviors, ranging from locomotion and opponent avoidance to acrobatics and combat tasks.
Figure~\ref{fig:tasks} illustrates examples of behaviors for various downstream tasks.

As summarized in Table~\ref{tab:downstream_task_return} and Figure~\ref{fig:hlc_learning_curve}, DSD maintains strong performance across all six tasks. It achieves the highest mean returns on Backflip, Location, and Shield Bash, while remaining competitive on Steer, Dodging, and Beat Saber. ASE-APS is the closest baseline overall, but exhibits larger gaps on Backflip and Shield Bash. ASE performs strongly on Steer and Dodging but is less competitive on other tasks, while ASE-DADS and ASE-METRA show substantially more uneven performance across tasks. Overall, the results show that DSD supports effective reuse across a broad range of downstream tasks. 

The relatively large variation in downstream task returns across independently trained models on Backflip and Shield Bash may reflect whether the pretrained policy can produce the behaviors required by these tasks, since the high-level controller cannot invoke a behavior absent from the pretrained low-level policy. We therefore examine the behaviors in each pretrained model. In Backflip, four out of five DSD policies produce successful backflips, compared with three from ASE-APS, one each from ASE-DADS and ASE-METRA, and none from ASE. The four successful DSD policies achieve a return of $0.85 \pm 0.01$, whereas the remaining policy obtains $0.35$. A similar pattern appears on Shield Bash: four of five DSD policies successfully perform the task and achieve $0.45 \pm 0.01$, whereas the remaining unsuccessful policy obtains $0.17$. In both tasks, the variation in downstream performance across independently trained models corresponds to whether the pretrained low-level policy can produce the behavior required by the task. Detailed success criteria for Backflip and Shield Bash are provided in Appendix~\ref{app:behavior_analysis}.

\subsection{Zero-shot Control Performance}
Zero-shot control selects task-relevant behaviors directly from an offline pool of trajectories generated by the pretrained skill policy, without additional policy training. For each task, candidate trajectories are evaluated using a task-specific fitness function, and the highest-fitness trajectory is then selected. Figure~\ref{fig:zeroshot_tasks} shows selected behaviors across tasks ranging from locomotion to combat.

We evaluate models pretrained on the Reallusion dataset by randomizing the goal for each task and recording the highest fitness score achieved within the offline trajectory pool. As shown in Table~\ref{tab:zeroshot_results}, DSD achieves the highest mean performance on eight of the ten tasks, spanning locomotion, full-body positioning, avoidance, and combat interactions. The largest gains appear on Jump and Duck. On these tasks, DSD also exhibits lower variation across independently trained models, suggesting that the corresponding behaviors are discovered more reliably across training runs.

\begin{figure}[t]
    \centering
    \begin{subfigure}{0.95\linewidth}
        \centering
        \includegraphics[width=\linewidth]{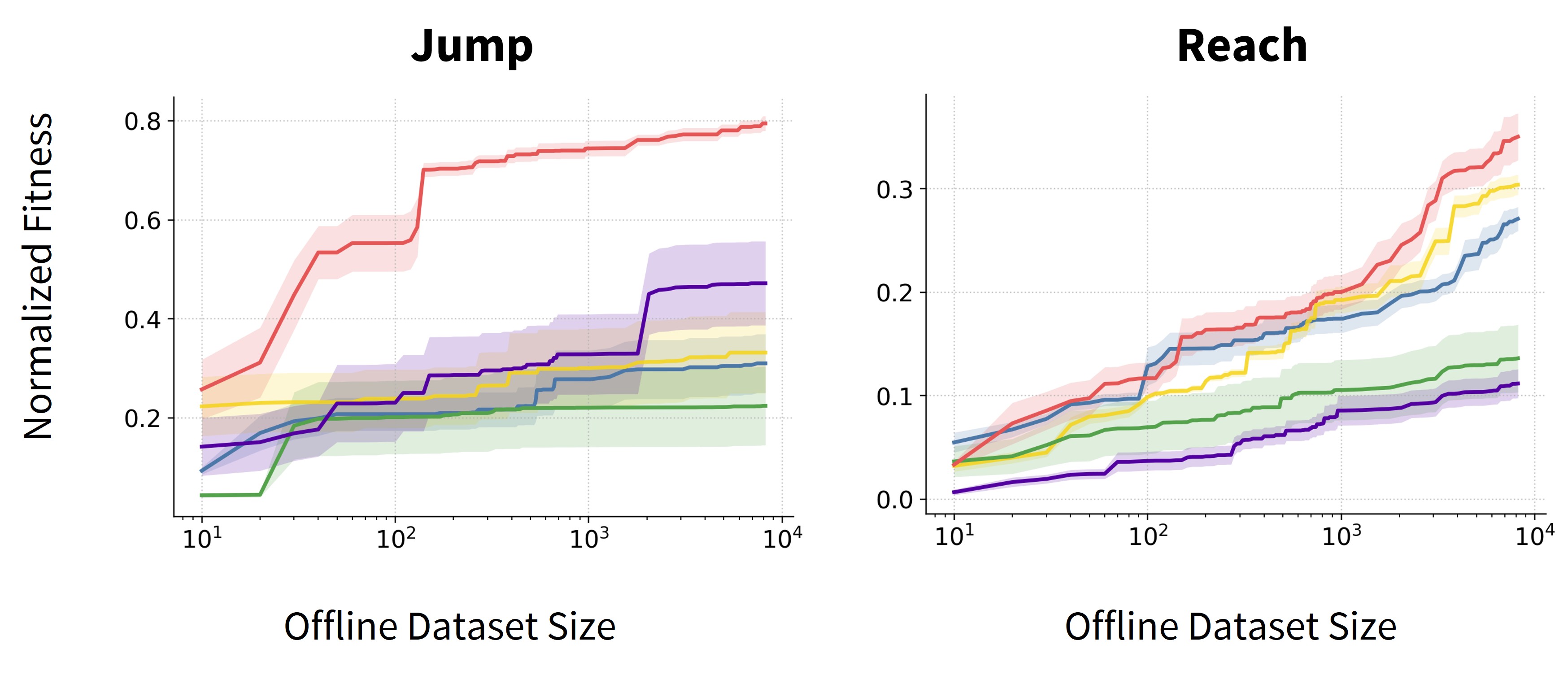}
    \end{subfigure}
    \begin{subfigure}{0.95\linewidth}
        \centering
        \includegraphics[width=\linewidth]{fig/zeroshot_datasize_legend.jpg}
    \end{subfigure}    
    \caption{Zero-shot performance as a function of offline trajectory pool size on Jump and Reach. The x-axis shows the number of offline trajectories on a log scale, and the y-axis reports the highest fitness retrieved from each pool. DSD achieves higher fitness across most pool sizes, indicating that task-suitable trajectories can be retrieved from smaller pools.}
    \label{fig:zeroshot_dataset_test}
\end{figure}

The performance of zero-shot control depends on whether task-relevant behaviors appear within a finite pool of offline trajectories. A more diverse skill set should increase the likelihood that this pool contains a trajectory suitable for a given task. To examine this property, we vary the size of the offline trajectory pool for Jump and Reach. Figure~\ref{fig:zeroshot_dataset_test} reports the fitness of the best candidate available at each pool size. DSD achieves higher fitness with fewer candidate trajectories on both tasks and maintains its advantage as the pool grows. This indicates that task-relevant behaviors are more readily available in the pools generated by DSD.

\begin{table}[t]
\caption{
Task returns for downstream control tasks using different pretrained skill models. For each method, five high-level controllers are trained separately using five independently trained low-level policies, with 4,096 evaluation episodes per model. DSD achieves strong mean performance across tasks with skill requirements.
}
\centering
\small
\setlength{\tabcolsep}{2pt}
\resizebox{\linewidth}{!}{%
\begin{tabular}{c|c|ccccc}
\toprule
\multirow{2}{*}{\textbf{Character}}
&
\multirow{2}{*}{\textbf{Task}}
&
\multicolumn{5}{c}{\textbf{Task Return}}
\\
&
&
{ASE}
&
\begin{tabular}{c}
{ASE-}\\[-0.5ex]
{APS}
\end{tabular}
&
\begin{tabular}{c}
{ASE-}\\[-0.5ex]
{DADS}
\end{tabular}
&
\begin{tabular}{c}
{ASE-}\\[-0.5ex]
{METRA}
\end{tabular}
&
\begin{tabular}{c}
{DSD}\\[-0.5ex]
{(Ours)}
\end{tabular}
\\
\midrule

\multirow{2}{*}{Humanoid}
& Steer
& \cellcolor{gray!20}{$0.85^{\pm 0.01}$} & $0.84^{\pm 0.02}$ & $0.65^{\pm 0.16}$ & $0.69^{\pm 0.02}$ & $0.83^{\pm 0.02}$
\\

& Backflip
& $0.44^{\pm 0.17}$ & $0.65^{\pm 0.25}$ & $0.41^{\pm 0.19}$ & $0.45^{\pm 0.21}$ & \cellcolor{gray!20}{$0.75^{\pm 0.20}$}
\\

\midrule

\multirow{4}{*}{%
\begin{tabular}{c}
Sword\\[-0.5ex]
\&\\[-0.5ex]
Shield
\end{tabular}}
& Location
& $0.42^{\pm 0.05}$ & $0.46^{\pm 0.06}$ & $0.46^{\pm 0.11}$ & $0.36^{\pm 0.01}$ & \cellcolor{gray!20}{$0.48^{\pm 0.01}$}
\\

& Dodging
& \cellcolor{gray!20}{$0.46^{\pm 0.04}$} & $0.44^{\pm 0.00}$ & $0.39^{\pm 0.13}$ & $0.27^{\pm 0.04}$ & $0.41^{\pm 0.00}$
\\

& \begin{tabular}{c}
Shield Bash
\end{tabular}
& $0.36^{\pm 0.19}$ & $0.31^{\pm 0.17}$ & $0.20^{\pm 0.12}$ & $0.08^{\pm 0.00}$ & \cellcolor{gray!20}{$0.39^{\pm 0.11}$}
\\

& \begin{tabular}{c}
Beat Saber
\end{tabular}
& $0.48^{\pm 0.06}$ & \cellcolor{gray!20}{$0.64^{\pm 0.05}$} & $0.51^{\pm 0.18}$ & $0.19^{\pm 0.04}$ & $0.61^{\pm 0.02}$
\\

\bottomrule
\end{tabular}%
}
\label{tab:downstream_task_return}
\end{table}

\begin{table*}[t]
\centering
\setlength{\tabcolsep}{3pt}
\caption{Zero-shot downstream performance measured from offline trajectory pools. Results are computed over five independently trained models, with a separate trajectory pool collected from each model. For each task, the goal condition is randomized and the highest fitness achieved by the selected trajectory is reported. Best results are highlighted in gray, and second-best results are underlined. DSD achieves the highest mean fitness on eight of the ten tasks.}
\label{tab:zeroshot_results}

\begin{tabular}{lcccccccccc}
\toprule

\textbf{Method}
& \textbf{Strafe}
& \textbf{Run}
& \textbf{Location}
& \textbf{Reach}
& \textbf{Jump}
& \textbf{Duck}
& \textbf{Dodge}
& \textbf{Sword Attack}
& \textbf{Shield Attack}
& \textbf{Kick Attack}
 \\

\midrule

\textbf{ASE}
& \underline{$0.81^{\pm 0.02}$}
& \cellcolor{gray!20}{$0.64^{\pm 0.03}$}
& $0.71^{\pm 0.03}$
& $0.25^{\pm 0.09}$
& $0.31^{\pm 0.26}$
& $0.37^{\pm 0.34}$
& $0.55^{\pm 0.01}$
& \underline{$0.63^{\pm 0.02}$}
& $0.47^{\pm 0.02}$
& \underline{$0.53^{\pm 0.02}$}
 \\

\textbf{ASE-APS}
& \underline{$0.81^{\pm 0.01}$}
& \cellcolor{gray!20}{$0.64^{\pm 0.01}$}
& \cellcolor{gray!20}{$0.74^{\pm 0.03}$}
& \underline{$0.31^{\pm 0.05}$}
& $0.33^{\pm 0.37}$
& $0.33^{\pm 0.34}$
& \underline{$0.60^{\pm 0.04}$}
& $0.62^{\pm 0.04}$ 
& \underline{$0.49^{\pm 0.02}$}
& \underline{$0.53^{\pm 0.02}$}
\\

\textbf{ASE-DADS}
& $0.74^{\pm 0.05}$
& $0.38^{\pm 0.06}$
& $0.58^{\pm 0.12}$
& $0.08^{\pm 0.06}$
& $0.23^{\pm 0.35}$
& $0.31^{\pm 0.36}$
& $0.23^{\pm 0.23}$
& $0.30^{\pm 0.27}$
& $0.26^{\pm 0.11}$
& $0.16^{\pm 0.14}$
 \\

\textbf{ASE-METRA}
& $0.69^{\pm 0.05}$
& $0.55^{\pm 0.02}$
& $0.49^{\pm 0.06}$
& $0.11^{\pm 0.06}$
& \underline{$0.47^{\pm 0.38}$}
& \underline{$0.56^{\pm 0.50}$}
& $0.33^{\pm 0.20}$
& $0.48^{\pm 0.14}$ 
& $0.31^{\pm 0.11}$
& $0.47^{\pm 0.12}$
\\

\midrule

\textbf{DSD (Ours)}
& \cellcolor{gray!20}{$0.82^{\pm 0.01}$}
& \underline{$0.63^{\pm 0.00}$}
& \underline{$0.73^{\pm 0.03}$}
& \cellcolor{gray!20}{$0.33^{\pm 0.11}$}
& \cellcolor{gray!20}{$0.81^{\pm 0.07}$}
& \cellcolor{gray!20}{$0.94^{\pm 0.02}$}
& \cellcolor{gray!20}{$0.62^{\pm 0.03}$}
& \cellcolor{gray!20}{$0.66^{\pm 0.02}$}
& \cellcolor{gray!20}{$0.51^{\pm 0.01}$}
& \cellcolor{gray!20}{$0.55^{\pm 0.02}$}
 \\

\bottomrule
\end{tabular}
\end{table*}

\subsection{Ablation Analysis}
To evaluate the impact of key design decisions for DSD, we conduct a series of ablation experiments to isolate the effects of individual components.

\paragraph{JIDS Ablation.}
JIDS is designed to prevent high-frequency motion artifacts from being rewarded as behavioral diversity. We therefore evaluate whether JIDS reduces such artifacts while preserving the diversity of the learned skills. We compare the full DSD model with a variant that provides unfiltered observations to the diffusion model and skill encoder. Motion smoothness is measured using the jerk of the root and joints. We additionally report FID and diversity to examine whether filtering improves motion quality without reducing behavioral diversity. As shown in Table~\ref{tab:lpf-ablation}, JIDS reduces root and joint jerk by approximately $13\%$ and $10\%$, respectively, indicating fewer jittering artifacts. JIDS also reduces FID by approximately $12\%$, suggesting improved overall motion quality. Meanwhile, Div-R increases by $3.2\%$, while Div-J decreases by $2.3\%$. These results show that JIDS suppresses high-frequency jitter and improves motion quality while largely preserving the diversity of the learned skills.

\paragraph{Conditional Entropy Ablation}
For a latent skill representation to be reusable, selecting the same latent should reliably produce a consistent behavior. Marginal state-entropy maximization encourages broad state coverage, but does not by itself enforce this consistency within each latent. To examine the role of conditional entropy, we compare the full DSD objective with a marginal-only variant that omits the conditional entropy reward, using models trained on the Reallusion dataset. 

To evaluate whether trajectories generated with a fixed latent remain behaviorally consistent, we consider the fixed-latent motion entropy (FME). FME measures how often a trajectory generated by a single latent produces behaviors that resemble different reference motion clips. For each trajectory $\tau^i$ generated with a fixed latent, every state transition $(\vc{s}_t^i,\vc{s}_{t+1}^i)$ is matched to its nearest reference motion clip. Let $p_i(m^j)$ denote the fraction of transitions matched to reference clip $m^j$. FME is defined as the entropy of this distribution:
\begin{equation}
\mathrm{FME}(\tau^i)
=
-\sum_j p_i(m^j)\log p_i(m^j).
\end{equation}
Lower FME indicates that a fixed-latent trajectory generates motions that match a fewer number of reference motion clips, which indicates more consistent behaviors from a given latent. Across five independently trained models, the marginal-only models yield an FME of $1.08 \pm 0.24$, compared with $0.60 \pm 0.02$ for DSD. The generated trajectories show the same pattern: DSD produces more consistent behaviors, whereas the marginal-only model often mixes a variety of different skills given the same latent.

To determine whether marginal state-entropy maximization alone is sufficient for downstream reuse, we additionally train task-specific high-level controllers using the pretrained marginal-only policies. As shown in Table~\ref{tab:down_ablation}, removing the conditional entropy reward consistently reduces downstream performance. Together, these results show that conditional entropy promotes consistent latent skills that can be more effectively reused for downstream tasks.

\paragraph{Motion Prior Ablation.}
Diverse skills alone do not guarantee that the learned behaviors are naturalistic. Next, we evaluate DSD without the motion-prior reward. Without the motion prior, DSD continues to discover diverse behaviors, but many of the resulting skills contain unnatural behaviors. In contrast, the full model discovers natural and recognizable skills, including backflips, kicks, and punches on MimicKit, as well as dancing, running, jumping, and various walking styles on LaFAN1.

The unnatural skills discovered without the motion prior can still be effective for downstream control. As shown in Table~\ref{tab:down_ablation}, removing the motion prior yields comparable or higher returns on the evaluated tasks, despite producing highly unnatural behaviors. Examples of these unnatural behaviors are shown in the supplementary video. This result shows that task return alone does not capture motion naturalness, highlighting the role of the motion prior in encouraging the skill discovery process toward more naturalistic behaviors.

\paragraph{Comparison with ExDM\@.}
The preceding ablation experiments examine the effects of conditional entropy, the motion prior, and JIDS individually. ExDM provides a complementary setting in which several of these elements are absent simultaneously: ExDM does not incorporate skill latents nor conditional entropy, and does not employ a motion prior or JIDS~\citep{ying2025exploratorydiffusion}. This limits flexible downstream reuse, since ExDM fine-tunes its policy for a fixed task goal, requiring separate optimization when the goal changes. Finally, the absence of a motion prior leads to highly unnatural motions. These results further illustrate the importance of the components introduced in DSD for learning effective and reusable skills.

\begin{table}[t]
\caption{
Ablation study on the contribution of the conditional entropy (CE) reward. 
Task returns are reported for downstream control tasks on the Reallusion dataset.
}
\centering
\small
\setlength{\tabcolsep}{6pt}
\begin{tabular}{c|ccc}
\toprule
\multirow{2}{*}{\textbf{Task}}
&
\multicolumn{3}{c}{\textbf{Task Return}}
\\
&
\textbf{w/o CE}
&
\textbf{w/o AMP}
&
\textbf{DSD (Ours)}
\\
\midrule
Location
& $0.38^{\pm 0.05}$
& \cellcolor{gray!20}{$0.62^{\pm 0.05}$}
& $0.48^{\pm 0.01}$
\\
Dodging
& $0.18^{\pm 0.04}$
& \cellcolor{gray!20}{$0.50^{\pm 0.04}$}
& $0.41^{\pm 0.00}$
\\
Shield Bash
& $0.08^{\pm 0.02}$
& $0.38^{\pm 0.14}$
& \cellcolor{gray!20}{$0.39^{\pm 0.11}$}
\\
Beat Saber
& $0.19^{\pm 0.03}$ 
& $0.60^{\pm 0.05}$
& \cellcolor{gray!20}{$0.61^{\pm 0.02}$}
\\
\bottomrule
\end{tabular}
\label{tab:down_ablation}
\end{table}

\section{Discussion}
In this work, we presented DSD, a diffusion-based skill discovery method for learning diverse and reusable motor skills for physics-based character control. Our results suggest that broad state coverage alone is insufficient for learning a useful skill repertoire: diverse behaviors must also be organized into consistent latent-conditioned skills. DSD combines diffusion-based marginal state-entropy optimization with a conditional entropy objective to promote this structure, while a motion prior encourages the discovered behaviors toward the reference motion distribution. The learned low-level policies enable skills to be reused for a diverse range of downstream tasks through both hierarchical and zero-shot control.

While DSD shows that diffusion-based entropy estimation can effectively support the discovery of diverse motor skills, some areas could benefit from further improvement. The current objective encourages consistent behavior under each latent, but does not explicitly organize the latent space according to semantic relationships between skills. Future work could investigate more structured latent representations that better facilitate interpolation, composition, or direct selection of related behaviors. In addition, the current zero-shot control setting selects skills from a finite offline trajectory pool, so performance depends on whether a suitable behavior has been recorded in the pool. More direct latent search or task-conditioned retrieval could allow better use of the learned latent space of skills without requiring a large set of pre-generated trajectories. 

More broadly, DSD suggests a possible future direction towards building large general-purpose motor control models. Instead of learning separate skills for narrow domains, future systems could learn a shared skill representation that can support a wide range of downstream control tasks. These general-purpose reusable models could enable physically simulated characters to be more easily controlled across tasks that require different forms of movements, interactions, and adaptations.

\begin{table}
\centering
\small
\setlength{\tabcolsep}{3pt}
\caption{Ablation study of Jitter-Induced Diversity Suppression (JIDS) on the Reallusion dataset. Results are averaged over five independently trained models. JIDS reduces root and joint jerk and improves FID while largely preserving the diversity of the learned repertoire.}
\label{tab:lpf-ablation}
\begin{tabular}{lccccc}
\toprule
\textbf{Method}
& \textbf{Jerk-R} $\downarrow$ & \textbf{Jerk-J} $\downarrow$
& \textbf{FID} $\downarrow$ & \textbf{Div-R} $\uparrow$ & \textbf{Div-J }$\uparrow$
 \\
\midrule
\textbf{w/o JIDS}
& $0.52^{\pm 0.01}$ & $0.82^{\pm 0.02}$
& $1.87^{\pm 0.13}$ & $0.126^{\pm 0.009}$ & \cellcolor{gray!20}{$0.173^{\pm 0.006}$}\\
\textbf{DSD (Ours)}&\cellcolor{gray!20}{$0.45^{\pm 0.01}$} & \cellcolor{gray!20}{$0.74^{\pm 0.02}$} 
& \cellcolor{gray!20}{$1.64^{\pm 0.10}$} & \cellcolor{gray!20}{$0.130^{\pm 0.003}$} & $0.169^{\pm 0.004}$
 \\
\bottomrule
\end{tabular}
\end{table}

\begin{acks}
This work was supported by NSERC via a Discovery Grant (RGPIN-2015-04843), and a Canada CIFAR AI Chair.
\end{acks}

\bibliographystyle{ACM-Reference-Format}
\bibliography{reference}

@article{coros2010Generalized,
author = {Coros, Stelian and Beaudoin, Philippe and van de Panne, Michiel},
title = {Generalized biped walking control},
year = {2010},
issue_date = {July 2010},
publisher = {Association for Computing Machinery},
address = {New York, NY, USA},
volume = {29},
number = {4},
issn = {0730-0301},
url = {https://doi.org/10.1145/1778765.1781156},
doi = {10.1145/1778765.1781156},
journal = {ACM Trans. Graph.},
month = jul,
articleno = {130},
numpages = {9}
}

@inproceedings{hodgins1995animating,
author = {Hodgins, Jessica K. and Wooten, Wayne L. and Brogan, David C. and O'Brien, James F.},
title = {Animating human athletics},
year = {1995},
isbn = {0897917014},
publisher = {Association for Computing Machinery},
address = {New York, NY, USA},
url = {https://doi.org/10.1145/218380.218414},
doi = {10.1145/218380.218414},
booktitle = {Proceedings of the 22nd Annual Conference on Computer Graphics and Interactive Techniques},
pages = {71–78},
numpages = {8},
series = {SIGGRAPH '95}
}

@inproceedings{raibert1991animation,
author = {Raibert, Marc H. and Hodgins, Jessica K.},
title = {Animation of dynamic legged locomotion},
year = {1991},
isbn = {0897914368},
publisher = {Association for Computing Machinery},
address = {New York, NY, USA},
url = {https://doi.org/10.1145/122718.122755},
doi = {10.1145/122718.122755},
booktitle = {Proceedings of the 18th Annual Conference on Computer Graphics and Interactive Techniques},
pages = {349–358},
numpages = {10},
series = {SIGGRAPH '91}
}

@article{yin2007simbicon,
author = {Yin, KangKang and Loken, Kevin and van de Panne, Michiel},
title = {SIMBICON: simple biped locomotion control},
year = {2007},
issue_date = {July 2007},
publisher = {Association for Computing Machinery},
address = {New York, NY, USA},
volume = {26},
number = {3},
issn = {0730-0301},
url = {https://doi.org/10.1145/1276377.1276509},
doi = {10.1145/1276377.1276509},
journal = {ACM Trans. Graph.},
month = jul,
pages = {105–es},
numpages = {10}
}

@inproceedings{zordan2002mocap,
author = {Zordan, Victor Brian and Hodgins, Jessica K.},
title = {Motion capture-driven simulations that hit and react},
year = {2002},
isbn = {1581135734},
publisher = {Association for Computing Machinery},
address = {New York, NY, USA},
url = {https://doi.org/10.1145/545261.545276},
doi = {10.1145/545261.545276},
booktitle = {Proceedings of the 2002 ACM SIGGRAPH/Eurographics Symposium on Computer Animation},
pages = {89–96},
numpages = {8},
location = {San Antonio, Texas},
series = {SCA '02}
}

@article{deLasa2010Feature,
author = {de Lasa, Martin and Mordatch, Igor and Hertzmann, Aaron},
title = {Feature-based locomotion controllers},
year = {2010},
issue_date = {July 2010},
publisher = {Association for Computing Machinery},
address = {New York, NY, USA},
volume = {29},
number = {4},
issn = {0730-0301},
url = {https://doi.org/10.1145/1778765.1781157},
doi = {10.1145/1778765.1781157},
journal = {ACM Trans. Graph.},
month = jul,
articleno = {131},
numpages = {10}
}

@InProceedings{levine132013guided,
  title = 	 {Guided Policy Search},
  author = 	 {Levine, Sergey and Koltun, Vladlen},
  booktitle = 	 {Proceedings of the 30th International Conference on Machine Learning},
  pages = 	 {1--9},
  year = 	 {2013},
  editor = 	 {Dasgupta, Sanjoy and McAllester, David},
  volume = 	 {28},
  number =       {3},
  series = 	 {Proceedings of Machine Learning Research},
  address = 	 {Atlanta, Georgia, USA},
  month = 	 {17--19 Jun},
  publisher =    {PMLR},
  url = 	 {https://proceedings.mlr.press/v28/levine13.html}
}

@article{mordatch2012cio,
author = {Mordatch, Igor and Todorov, Emanuel and Popovi\'{c}, Zoran},
title = {Discovery of complex behaviors through contact-invariant optimization},
year = {2012},
issue_date = {July 2012},
publisher = {Association for Computing Machinery},
address = {New York, NY, USA},
volume = {31},
number = {4},
issn = {0730-0301},
url = {https://doi.org/10.1145/2185520.2185539},
doi = {10.1145/2185520.2185539},
journal = {ACM Trans. Graph.},
month = jul,
articleno = {43},
numpages = {8}
}

@inproceedings{da2008simulation,
  title={Simulation of human motion data using short-horizon model-predictive control},
  author={Da Silva, Marco and Abe, Yeuhi and Popovi{\'c}, Jovan},
  booktitle={Computer Graphics Forum},
  volume={27},
  number={2},
  pages={371--380},
  year={2008},
  organization={Wiley Online Library}
}

@article{tan2014bicycle,
author = {Tan, Jie and Gu, Yuting and Liu, C. Karen and Turk, Greg},
title = {Learning bicycle stunts},
year = {2014},
issue_date = {July 2014},
publisher = {Association for Computing Machinery},
address = {New York, NY, USA},
volume = {33},
number = {4},
issn = {0730-0301},
url = {https://doi.org/10.1145/2601097.2601121},
doi = {10.1145/2601097.2601121},
journal = {ACM Trans. Graph.},
month = jul,
articleno = {50},
numpages = {12}
}

@inproceedings{Panne1994virtual,
 author = {Michiel van de Panne and Ryan Kim and Eugene Fiume},
 title = {Virtual Wind-up Toys for Animation},
 booktitle = {Proceedings of Graphics Interface '94},
 series = {GI '94},
 year = {1994},
 isbn = {0-9695338-3-7},
 issn = {0713-5424},
 location = {Banff, Alberta, Canada},
 pages = {208--215},
 numpages = {8},
 url = {http://graphicsinterface.org/wp-content/uploads/gi1994-25.pdf},
 publisher = {Canadian Human-Computer Communications Society},
 address = {Toronto, Ontario, Canada},
}

@article{geyer2003positive,
  title={Positive force feedback in bouncing gaits?},
  author={Geyer, Hartmut and Seyfarth, Andre and Blickhan, Reinhard},
  journal={Proceedings of the Royal Society of London. Series B: Biological Sciences},
  volume={270},
  number={1529},
  pages={2173--2183},
  year={2003},
  publisher={The Royal Society}
}

@article{ha2012falling,
author = {Ha, Sehoon and Ye, Yuting and Liu, C. Karen},
title = {Falling and landing motion control for character animation},
year = {2012},
issue_date = {November 2012},
publisher = {Association for Computing Machinery},
address = {New York, NY, USA},
volume = {31},
number = {6},
issn = {0730-0301},
url = {https://doi.org/10.1145/2366145.2366174},
doi = {10.1145/2366145.2366174},
journal = {ACM Trans. Graph.},
month = nov,
articleno = {155},
numpages = {9}
}

@article{gejitenbeek2013flexible,
author = {Geijtenbeek, Thomas and van de Panne, Michiel and van der Stappen, A. Frank},
title = {Flexible muscle-based locomotion for bipedal creatures},
year = {2013},
issue_date = {November 2013},
publisher = {Association for Computing Machinery},
address = {New York, NY, USA},
volume = {32},
number = {6},
issn = {0730-0301},
url = {https://doi.org/10.1145/2508363.2508399},
doi = {10.1145/2508363.2508399},
journal = {ACM Trans. Graph.},
month = nov,
articleno = {206},
numpages = {11}
}

@article{wang2012optimizing,
author = {Wang, Jack M. and Hamner, Samuel R. and Delp, Scott L. and Koltun, Vladlen},
title = {Optimizing locomotion controllers using biologically-based actuators and objectives},
year = {2012},
issue_date = {July 2012},
publisher = {Association for Computing Machinery},
address = {New York, NY, USA},
volume = {31},
number = {4},
issn = {0730-0301},
url = {https://doi.org/10.1145/2185520.2185521},
doi = {10.1145/2185520.2185521},
journal = {ACM Trans. Graph.},
month = jul,
articleno = {25},
numpages = {11}
}

@article{lee2009biomech,
author = {Lee, Sung-Hee and Sifakis, Eftychios and Terzopoulos, Demetri},
title = {Comprehensive biomechanical modeling and simulation of the upper body},
year = {2009},
issue_date = {August 2009},
publisher = {Association for Computing Machinery},
address = {New York, NY, USA},
volume = {28},
number = {4},
issn = {0730-0301},
url = {https://doi.org/10.1145/1559755.1559756},
doi = {10.1145/1559755.1559756},
journal = {ACM Trans. Graph.},
month = sep,
articleno = {99},
numpages = {17}
}

@article{kwon2017momentum,
author = {Kwon, Taesoo and Hodgins, Jessica K.},
title = {Momentum-Mapped Inverted Pendulum Models for Controlling Dynamic Human Motions},
year = {2017},
issue_date = {August 2017},
publisher = {Association for Computing Machinery},
address = {New York, NY, USA},
volume = {36},
number = {4},
issn = {0730-0301},
url = {https://doi.org/10.1145/3072959.2983616},
doi = {10.1145/3072959.2983616},
journal = {ACM Trans. Graph.},
month = jul,
articleno = {145d},
numpages = {14}
}

@article{lee2010biped,
author = {Lee, Yoonsang and Kim, Sungeun and Lee, Jehee},
title = {Data-driven biped control},
year = {2010},
issue_date = {July 2010},
publisher = {Association for Computing Machinery},
address = {New York, NY, USA},
volume = {29},
number = {4},
issn = {0730-0301},
url = {https://doi.org/10.1145/1778765.1781155},
doi = {10.1145/1778765.1781155},
journal = {ACM Trans. Graph.},
month = jul,
articleno = {129},
numpages = {8}
}

@inproceedings{sharon2005synthesis,
  title={Synthesis of controllers for stylized planar bipedal walking},
  author={Sharon, Dana and van de Panne, Michiel},
  booktitle={Proceedings of the 2005 IEEE International Conference on Robotics and Automation},
  pages={2387--2392},
  year={2005},
  organization={IEEE}
}

@article{liu2005learning,
author = {Liu, C. Karen and Hertzmann, Aaron and Popovi\'{c}, Zoran},
title = {Learning physics-based motion style with nonlinear inverse optimization},
year = {2005},
issue_date = {July 2005},
publisher = {Association for Computing Machinery},
address = {New York, NY, USA},
volume = {24},
number = {3},
issn = {0730-0301},
url = {https://doi.org/10.1145/1073204.1073314},
doi = {10.1145/1073204.1073314},
journal = {ACM Trans. Graph.},
month = jul,
pages = {1071–1081},
numpages = {11}
}

@article{sok200simulating,
author = {Sok, Kwang Won and Kim, Manmyung and Lee, Jehee},
title = {Simulating biped behaviors from human motion data},
year = {2007},
issue_date = {July 2007},
publisher = {Association for Computing Machinery},
address = {New York, NY, USA},
volume = {26},
number = {3},
issn = {0730-0301},
url = {https://doi.org/10.1145/1276377.1276511},
doi = {10.1145/1276377.1276511},
journal = {ACM Trans. Graph.},
month = jul,
pages = {107–es},
numpages = {10}
}

@article{liu2012terrain,
author = {Liu, Libin and Yin, KangKang and van de Panne, Michiel and Guo, Baining},
title = {Terrain runner: control, parameterization, composition, and planning for highly dynamic motions},
year = {2012},
issue_date = {November 2012},
publisher = {Association for Computing Machinery},
address = {New York, NY, USA},
volume = {31},
number = {6},
issn = {0730-0301},
url = {https://doi.org/10.1145/2366145.2366173},
doi = {10.1145/2366145.2366173},
journal = {ACM Trans. Graph.},
month = nov,
articleno = {154},
numpages = {10}
}

@article{liu2010sampling,
author = {Liu, Libin and Yin, KangKang and van de Panne, Michiel and Shao, Tianjia and Xu, Weiwei},
title = {Sampling-based contact-rich motion control},
year = {2010},
issue_date = {July 2010},
publisher = {Association for Computing Machinery},
address = {New York, NY, USA},
volume = {29},
number = {4},
issn = {0730-0301},
url = {https://doi.org/10.1145/1778765.1778865},
doi = {10.1145/1778765.1778865},
journal = {ACM Trans. Graph.},
month = jul,
articleno = {128},
numpages = {10}
}

@article{liu2016guided,
  title={Guided learning of control graphs for physics-based characters},
  author={Liu, Libin and Panne, Michiel Van De and Yin, KangKang},
  journal={ACM Transactions on Graphics (TOG)},
  volume={35},
  number={3},
  pages={1--14},
  year={2016},
  publisher={ACM New York, NY, USA}
}

@article{peng2018deepmimic,
	author = {Peng, Xue Bin and Abbeel, Pieter and Levine, Sergey and van de Panne, Michiel},
	title = {DeepMimic: Example-guided Deep Reinforcement Learning of Physics-based Character Skills},
	journal = {ACM Trans. Graph.},
	issue_date = {August 2018},
	volume = {37},
	number = {4},
	month = jul,
	year = {2018},
	issn = {0730-0301},
	pages = {143:1--143:14},
	articleno = {143},
	numpages = {14},
	url = {http://doi.acm.org/10.1145/3197517.3201311},
	doi = {10.1145/3197517.3201311},
	acmid = {3201311},
	publisher = {ACM},
	address = {New York, NY, USA},
}

@inproceedings{chentanez2018physics,
author = {Chentanez, Nuttapong and M\"{u}ller, Matthias and Macklin, Miles and Makoviychuk, Viktor and Jeschke, Stefan},
title = {Physics-based motion capture imitation with deep reinforcement learning},
year = {2018},
isbn = {9781450360159},
publisher = {Association for Computing Machinery},
address = {New York, NY, USA},
url = {https://doi.org/10.1145/3274247.3274506},
doi = {10.1145/3274247.3274506},
booktitle = {Proceedings of the 11th ACM SIGGRAPH Conference on Motion, Interaction and Games},
articleno = {1},
numpages = {10},
location = {Limassol, Cyprus},
series = {MIG '18}
}

@article{lee2019scalable,
author = {Lee, Seunghwan and Park, Moonseok and Lee, Kyoungmin and Lee, Jehee},
title = {Scalable muscle-actuated human simulation and control},
year = {2019},
issue_date = {August 2019},
publisher = {Association for Computing Machinery},
address = {New York, NY, USA},
volume = {38},
number = {4},
issn = {0730-0301},
url = {https://doi.org/10.1145/3306346.3322972},
doi = {10.1145/3306346.3322972},
journal = {ACM Trans. Graph.},
month = jul,
articleno = {73},
numpages = {13}
}

@inproceedings{coros2009robust,
author = {Coros, Stelian and Beaudoin, Philippe and van de Panne, Michiel},
title = {Robust task-based control policies for physics-based characters},
year = {2009},
isbn = {9781605588582},
publisher = {Association for Computing Machinery},
address = {New York, NY, USA},
url = {https://doi.org/10.1145/1661412.1618516},
doi = {10.1145/1661412.1618516},
booktitle = {ACM SIGGRAPH Asia 2009 Papers},
articleno = {170},
numpages = {9},
location = {Yokohama, Japan},
series = {SIGGRAPH Asia '09}
}

@inproceedings{hausman2018learning,
  title={Learning an embedding space for transferable robot skills},
  author={Hausman, Karol and Springenberg, Jost Tobias and Wang, Ziyu and Heess, Nicolas and Riedmiller, Martin},
  booktitle={International Conference on Learning Representations},
  year={2018}
}

@article{heess2016learning,
  title={Learning and transfer of modulated locomotor controllers},
  author={Heess, Nicolas and Wayne, Greg and Tassa, Yuval and Lillicrap, Timothy and Riedmiller, Martin and Silver, David},
  journal={arXiv preprint arXiv:1610.05182},
  year={2016}
}

@article{
	peng2021AMP,
	author = {Peng, Xue Bin and Ma, Ze and Abbeel, Pieter and Levine, Sergey and Kanazawa, Angjoo},
	title = {AMP: Adversarial Motion Priors for Stylized Physics-Based Character Control},
	journal = {ACM Trans. Graph.},
	issue_date = {August 2021},
	volume = {40},
	number = {4},
	month = jul,
	year = {2021},
	articleno = {1},
	numpages = {15},
	url = {http://doi.acm.org/10.1145/3450626.3459670},
	doi = {10.1145/3450626.3459670},
	publisher = {ACM},
	address = {New York, NY, USA},
}

@inproceedings{
    zhang2025ADD,
    author={Zhang, Ziyu and Bashkirov, Sergey and Yang, Dun and Shi, Yi and Taylor, Michael and Peng, Xue Bin},
    title = {Physics-Based Motion Imitation with Adversarial Differential Discriminators},
    year = {2025},
    booktitle = {SIGGRAPH Asia 2025 Conference Papers (SIGGRAPH Asia '25 Conference Papers)}
}

@article{burda2018exploration,
  title={Exploration by random network distillation},
  author={Burda, Yuri and Edwards, Harrison and Storkey, Amos and Klimov, Oleg},
  journal={arXiv preprint arXiv:1810.12894},
  year={2018}
}

@inproceedings{pathak2017exploration,
author = {Pathak, Deepak and Agrawal, Pulkit and Efros, Alexei A. and Darrell, Trevor},
title = {Curiosity-driven exploration by self-supervised prediction},
year = {2017},
publisher = {JMLR.org},
booktitle = {Proceedings of the 34th International Conference on Machine Learning - Volume 70},
pages = {2778–2787},
numpages = {10},
location = {Sydney, NSW, Australia},
series = {ICML'17}
}

@inproceedings{
hansen2020Fast,
title={Fast Task Inference with Variational Intrinsic Successor Features},
author={Steven Hansen and Will Dabney and Andre Barreto and David Warde-Farley and Tom Van de Wiele and Volodymyr Mnih},
booktitle={International Conference on Learning Representations},
year={2020},
url={https://openreview.net/forum?id=BJeAHkrYDS}
}

@inproceedings{bellemare2016unifying,
author = {Bellemare, Marc G. and Srinivasan, Sriram and Ostrovski, Georg and Schaul, Tom and Saxton, David and Munos, R\'{e}mi},
title = {Unifying count-based exploration and intrinsic motivation},
year = {2016},
isbn = {9781510838819},
publisher = {Curran Associates Inc.},
address = {Red Hook, NY, USA},
booktitle = {Proceedings of the 30th International Conference on Neural Information Processing Systems},
pages = {1479–1487},
numpages = {9},
location = {Barcelona, Spain},
series = {NIPS'16}
}

@inproceedings{tang2017exploration,
author = {Tang, Haoran and Houthooft, Rein and Foote, Davis and Stooke, Adam and Chen, Xi and Duan, Yan and Schulman, John and De Turck, Filip and Abbeel, Pieter},
title = {\#Exploration: a study of count-based exploration for deep reinforcement learning},
year = {2017},
isbn = {9781510860964},
publisher = {Curran Associates Inc.},
address = {Red Hook, NY, USA},
booktitle = {Proceedings of the 31st International Conference on Neural Information Processing Systems},
pages = {2750–2759},
numpages = {10},
location = {Long Beach, California, USA},
series = {NIPS'17}
}

@inproceedings{hazan2019provably,
  title={Provably efficient maximum entropy exploration},
  author={Hazan, Elad and Kakade, Sham and Singh, Karan and Van Soest, Abby},
  booktitle={International Conference on Machine Learning},
  pages={2681--2691},
  year={2019},
  organization={PMLR}
}

@inproceedings{
liu2021behavior,
title={Behavior From the Void: Unsupervised Active Pre-Training},
author={Hao Liu and Pieter Abbeel},
booktitle={Advances in Neural Information Processing Systems},
editor={A. Beygelzimer and Y. Dauphin and P. Liang and J. Wortman Vaughan},
year={2021},
url={https://openreview.net/forum?id=fIn4wLS2XzU}
}

@misc{ying2025exploratorydiffusion,
      title={Exploratory Diffusion Model for Unsupervised Reinforcement Learning}, 
      author={Chengyang Ying and Huayu Chen and Xinning Zhou and Zhongkai Hao and Hang Su and Jun Zhu},
      year={2025},
      eprint={2502.07279},
      archivePrefix={arXiv},
      primaryClass={cs.LG},
      url={https://arxiv.org/abs/2502.07279}, 
}

@inproceedings{park2021lsd,
  title={Lipschitz-constrained unsupervised skill discovery},
  author={Park, Seohong and Choi, Jongwook and Kim, Jaekyeom and Lee, Honglak and Kim, Gunhee},
  booktitle={International Conference on Learning Representations},
  year={2021}
}

@inproceedings{
park2024metra,
title={{METRA}: Scalable Unsupervised {RL} with Metric-Aware Abstraction},
author={Seohong Park and Oleh Rybkin and Sergey Levine},
booktitle={The Twelfth International Conference on Learning Representations},
year={2024},
url={https://openreview.net/forum?id=c5pwL0Soay}
}

@misc{eysenbach2018diversity,
      title={Diversity is All You Need: Learning Skills without a Reward Function}, 
      author={Benjamin Eysenbach and Abhishek Gupta and Julian Ibarz and Sergey Levine},
      year={2018},
      eprint={1802.06070},
      archivePrefix={arXiv},
      primaryClass={cs.AI},
      url={https://arxiv.org/abs/1802.06070}, 
}

@article{sharma2019dynamics,
  title={Dynamics-aware unsupervised discovery of skills},
  author={Sharma, Archit and Gu, Shixiang and Levine, Sergey and Kumar, Vikash and Hausman, Karol},
  journal={arXiv preprint arXiv:1907.01657},
  year={2019}
}

@inproceedings{liu2021aps,
  title={Aps: Active pretraining with successor features},
  author={Liu, Hao and Abbeel, Pieter},
  booktitle={International Conference on Machine Learning},
  pages={6736--6747},
  year={2021},
  organization={PMLR}
}

@article{gregor2016variational,
  title={Variational intrinsic control},
  author={Gregor, Karol and Rezende, Danilo Jimenez and Wierstra, Daan},
  journal={arXiv preprint arXiv:1611.07507},
  year={2016}
}

@inproceedings{baumli2021relative,
  title={Relative variational intrinsic control},
  author={Baumli, Kate and Warde-Farley, David and Hansen, Steven and Mnih, Volodymyr},
  booktitle={Proceedings of the AAAI conference on artificial intelligence},
  volume={35},
  number={8},
  pages={6732--6740},
  year={2021}
}

@article{tirinzoni2025zero,
  title={Zero-shot whole-body humanoid control via behavioral foundation models},
  author={Tirinzoni, Andrea and Touati, Ahmed and Farebrother, Jesse and Guzek, Mateusz and Kanervisto, Anssi and Xu, Yingchen and Lazaric, Alessandro and Pirotta, Matteo},
  journal={arXiv preprint arXiv:2504.11054},
  year={2025}
}

@article{peng2022ASE,
	author = {Peng, Xue Bin and Guo, Yunrong and Halper, Lina and Levine, Sergey and Fidler, Sanja},
	title = {ASE: Large-scale Reusable Adversarial Skill Embeddings for Physically Simulated Characters},
	journal = {ACM Trans. Graph.},
	issue_date = {August 2022},
	volume = {41},
	number = {4},
	month = jul,
	year = {2022},
	articleno = {94},
	publisher = {ACM},
	address = {New York, NY, USA}
}

@inproceedings{
	tessler2023calm,
	author={Tessler, Chen and Kasten, Yoni and Guo, Yunrong and Mannor, Shie and Chechik, Gal and Peng, Xue Bin},
	title = {CALM: Conditional Adversarial Latent Models for Directable Virtual Characters},
	year = {2023},
	isbn = {9798400701597},
	publisher = {Association for Computing Machinery},
	address = {New York, NY, USA},
	url = {https://doi.org/10.1145/3588432.3591541},
	doi = {10.1145/3588432.3591541},
	booktitle = {ACM SIGGRAPH 2023 Conference Proceedings},
	location = {Los Angeles, CA, USA},
	series = {SIGGRAPH '23}
}

@article{dou2022case,
  title={C·ASE: Learning Conditional Adversarial Skill Embeddings for Physics-based Characters},
  author={Zhiyang Dou and Xuelin Chen and Qingnan Fan and Taku Komura and Wenping Wang},
  journal={arXiv preprint arXiv:2309.11351},
  year={2023}
}

@book{sutton1998rl,
  author={Sutton, R.S. and Barto, A.G.},
  journal={IEEE Transactions on Neural Networks}, 
  title={Reinforcement Learning: An Introduction}, 
  year={1998},
  volume={9},
  number={5},
  pages={1054-1054},
  doi={10.1109/TNN.1998.712192}}

@inproceedings{ho2020ddpm,
author = {Ho, Jonathan and Jain, Ajay and Abbeel, Pieter},
title = {Denoising diffusion probabilistic models},
year = {2020},
isbn = {9781713829546},
publisher = {Curran Associates Inc.},
address = {Red Hook, NY, USA},
booktitle = {Proceedings of the 34th International Conference on Neural Information Processing Systems},
articleno = {574},
numpages = {12},
location = {Vancouver, BC, Canada},
series = {NIPS '20}
}

@inproceedings{wang2023prolific,
author = {Wang, Zhengyi and Lu, Cheng and Wang, Yikai and Bao, Fan and Li, Chongxuan and Su, Hang and Zhu, Jun},
title = {ProlificDreamer: high-fidelity and diverse text-to-3D generation with variational score distillation},
year = {2023},
publisher = {Curran Associates Inc.},
address = {Red Hook, NY, USA},
booktitle = {Proceedings of the 37th International Conference on Neural Information Processing Systems},
articleno = {368},
numpages = {36},
location = {New Orleans, LA, USA},
series = {NIPS '23}
}

@misc{mu2025smpreusablescorematchingmotion,
      title={SMP: Reusable Score-Matching Motion Priors for Physics-Based Character Control}, 
      author={Yuxuan Mu and Ziyu Zhang and Yi Shi and Minami Matsumoto and Kotaro Imamura and Guy Tevet and Chuan Guo and Michael Taylor and Chang Shu and Pengcheng Xi and Xue Bin Peng},
      year={2025},
      eprint={2512.03028},
      archivePrefix={arXiv},
      primaryClass={cs.GR},
      url={https://arxiv.org/abs/2512.03028}, 
}

@inproceedings{
poole2023dreamfusion,
title={DreamFusion: Text-to-3D using 2D Diffusion},
author={Ben Poole and Ajay Jain and Jonathan T. Barron and Ben Mildenhall},
booktitle={The Eleventh International Conference on Learning Representations },
year={2023},
url={https://openreview.net/forum?id=FjNys5c7VyY}
}

@article{wu2024diffusing,
  title={Diffusing States and Matching Scores: A New Framework for Imitation Learning},
  author={Wu, Runzhe and Chen, Yiding and Swamy, Gokul and Brantley, Kiant{\'e} and Sun, Wen},
  journal={arXiv preprint arXiv:2410.13855},
  year={2024}
}

@inproceedings{
tevet2023human,
title={Human Motion Diffusion Model},
author={Guy Tevet and Sigal Raab and Brian Gordon and Yoni Shafir and Daniel Cohen-Or and Amit Haim Bermano},
booktitle={The Eleventh International Conference on Learning Representations },
year={2023},
url={https://openreview.net/forum?id=SJ1kSyO2jwu}
}

@inproceedings{
jiang2024animated,
title={Animate3D: Animating Any 3D Model with Multi-view Video Diffusion},
author={Yanqin Jiang and Chaohui Yu and Chenjie Cao and Fan Wang and Weiming Hu and Jin Gao},
booktitle={The Thirty-eighth Annual Conference on Neural Information Processing Systems},
year={2024},
url={https://openreview.net/forum?id=HB6KaCFiMN}
}

@inproceedings{yin2024onestep,
      title={One-step Diffusion with Distribution Matching Distillation},
      author={Yin, Tianwei and Gharbi, Micha{\"e}l and Zhang, Richard and Shechtman, Eli and Durand, Fr{\'e}do and Freeman, William T and Park, Taesung},
      booktitle={CVPR},
      year={2024}
    }

@inproceedings{luo2024text,
      author={Luo, Calvin and He, Mandy and Zeng, Zilai and Sun, Chen},
      booktitle={Advances in Neural Information Processing Systems},
      title={Text-Aware Diffusion for Policy Learning},
      volume={37}, 
      year={2024}
     }

@inproceedings{kim2023FLAME,
author = {Kim, Jihoon and Kim, Jiseob and Choi, Sungjoon},
title = {FLAME: free-form language-based motion synthesis \& editing},
year = {2023},
isbn = {978-1-57735-880-0},
publisher = {AAAI Press},
url = {https://doi.org/10.1609/aaai.v37i7.25996},
doi = {10.1609/aaai.v37i7.25996},
booktitle = {Proceedings of the Thirty-Seventh AAAI Conference on Artificial Intelligence and Thirty-Fifth Conference on Innovative Applications of Artificial Intelligence and Thirteenth Symposium on Educational Advances in Artificial Intelligence},
articleno = {927},
numpages = {9},
series = {AAAI'23/IAAI'23/EAAI'23}
}

@inproceedings{cohan2024flexible,
author = {Cohan, Setareh and Tevet, Guy and Reda, Daniele and Peng, Xue Bin and van de Panne, Michiel},
title = {Flexible Motion In-betweening with Diffusion Models},
year = {2024},
isbn = {9798400705250},
publisher = {Association for Computing Machinery},
address = {New York, NY, USA},
url = {https://doi.org/10.1145/3641519.3657414},
doi = {10.1145/3641519.3657414},
booktitle = {ACM SIGGRAPH 2024 Conference Papers},
articleno = {69},
numpages = {9},
location = {Denver, CO, USA},
series = {SIGGRAPH '24}
}

@inproceedings{campos20exp,
author = {Campos, V\'{\i}ctor and Trott, Alex and Xiong, Caiming and Socher, Richard and Giro-i-Nieto, Xavier and Torres, Jordi},
title = {Explore, discover and learn: unsupervised discovery of state-covering skills},
year = {2020},
booktitle = {Proceedings of the 37th International Conference on Machine Learning},
articleno = {123},
numpages = {11},
series = {ICML'20}
}

@misc{schulman2017proximalpolicyoptimizationalgorithms,
      title={Proximal Policy Optimization Algorithms}, 
      author={John Schulman and Filip Wolski and Prafulla Dhariwal and Alec Radford and Oleg Klimov},
      year={2017},
      eprint={1707.06347},
      archivePrefix={arXiv},
      primaryClass={cs.LG},
      url={https://arxiv.org/abs/1707.06347}, 
}

@article{park2026periodic,
    title={Periodic skill discovery},
    author={Park, Jonghae and Cho, Daesol and Lee, Jusuk and Shim, Dongseok and Jang, Inkyu and Kim, H Jin},
    journal={Advances in Neural Information Processing Systems},
    volume={38},
    pages={105404--105438},
    year={2026}
  }

@inproceedings{cathomen2025d3,
  author    = {Cathomen, Rafael and Mittal, Mayank and Vlastelica, Marin and Hutter, Marco},
  title     = {Divide, Discover, Deploy: Factorized Skill Learning with Symmetry and Style Priors},
  booktitle = {Conference on Robot Learning (CoRL)},
  year      = {2025},
}

@misc{atanassov2024constrained,
      title={Constrained Skill Discovery: Quadruped Locomotion with Unsupervised Reinforcement Learning}, 
      author={Vassil Atanassov and Wanming Yu and Alexander Luis Mitchell and Mark Nicholas Finean and Ioannis Havoutis},
      year={2024},
      eprint={2410.07877},
      archivePrefix={arXiv},
      primaryClass={cs.RO},
      url={https://arxiv.org/abs/2410.07877}, 
}

@article{
      MimicKitPeng2025,
      title={MimicKit: A Reinforcement Learning Framework for Motion Imitation and Control}, 
      author={Peng, Xue Bin},
      year={2025},
      eprint={2510.13794},
      archivePrefix={arXiv},
      primaryClass={cs.GR},
      url={https://arxiv.org/abs/2510.13794}, 
}

@article{harvey20robust,
author = {Harvey, F\'{e}lix G. and Yurick, Mike and Nowrouzezahrai, Derek and Pal, Christopher},
title = {Robust motion in-betweening},
year = {2020},
issue_date = {August 2020},
publisher = {Association for Computing Machinery},
address = {New York, NY, USA},
volume = {39},
number = {4},
doi = {10.1145/3386569.3392480},
journal = {ACM Trans. Graph.},
month = aug,
articleno = {60},
numpages = {12},
}

@misc{reallusion2022,
  author       = {Reallusion},
  title        = {3D Animation and 2D Cartoons Made Simple},
  year         = {2022},
  url = {http://www.reallusion.com},
}

@misc{makoviychuk2021isaacgymhighperformance,
      title={Isaac Gym: High Performance GPU-Based Physics Simulation For Robot Learning}, 
      author={Viktor Makoviychuk and Lukasz Wawrzyniak and Yunrong Guo and Michelle Lu and Kier Storey and Miles Macklin and David Hoeller and Nikita Rudin and Arthur Allshire and Ankur Handa and Gavriel State},
      year={2021},
      eprint={2108.10470},
      archivePrefix={arXiv},
      primaryClass={cs.RO},
      url={https://arxiv.org/abs/2108.10470}, 
}

@INPROCEEDINGS{petrovich21Actor,
  author={Petrovich, Mathis and Black, Michael J. and Varol, Gül},
  booktitle={2021 IEEE/CVF International Conference on Computer Vision (ICCV)}, 
  title={Action-Conditioned 3D Human Motion Synthesis with Transformer VAE}, 
  year={2021},
  pages={10965-10975},
  doi={10.1109/ICCV48922.2021.01080}}

@article{touati2021learning,
  title={Learning one representation to optimize all rewards},
  author={Touati, Ahmed and Ollivier, Yann},
  journal={Advances in Neural Information Processing Systems},
  volume={34},
  pages={13--23},
  year={2021}
}

@article{frans2024unsupervised,
  title={Unsupervised zero-shot reinforcement learning via functional reward encodings},
  author={Frans, Kevin and Park, Seohong and Abbeel, Pieter and Levine, Sergey},
  journal={arXiv preprint arXiv:2402.17135},
  year={2024}
}

@misc{zhou2020continuityrotationrepresentationsneural,
      title={On the Continuity of Rotation Representations in Neural Networks}, 
      author={Yi Zhou and Connelly Barnes and Jingwan Lu and Jimei Yang and Hao Li},
      year={2020},
      eprint={1812.07035},
      archivePrefix={arXiv},
      primaryClass={cs.LG},
      url={https://arxiv.org/abs/1812.07035}, 
}

@article{grassia1998practical,
  title={Practical parameterization of rotations using the exponential map},
  author={Grassia, F Sebastian},
  journal={Journal of graphics tools},
  volume={3},
  number={3},
  pages={29--48},
  year={1998},
  publisher={Taylor \& Francis}
}

@article{schulman2015high,
  title={High-dimensional continuous control using generalized advantage estimation},
  author={Schulman, John and Moritz, Philipp and Levine, Sergey and Jordan, Michael and Abbeel, Pieter},
  journal={arXiv preprint arXiv:1506.02438},
  year={2015}
}

@inproceedings{kingma2015adam,
  title={Adam: A Method for Stochastic Optimization},
  author={Kingma, Diederik P and Ba, Jimmy},
  booktitle={3rd International Conference for Learning Representations (ICLR)},
  year={2015},
  url={https://arxiv.org/abs/1412.6980}
}

\clearpage
\appendix
\section*{Appendix}
\section{Experimental Setup}
\label{app:experimental setup}
All agents are simulated in Isaac Gym at 120\,Hz with GPU acceleration~\cite{makoviychuk2021isaacgymhighperformance}. During low-level policy training, 4096 environments run in parallel on a single NVIDIA H100 GPU, and the agent is controlled at 30\,Hz through PD targets. The low-level policy is trained with approximately 10 billion samples. For hierarchical control, the high-level policy operates at 6\,Hz. For zero-shot control, offline trajectories are collected from 8192 environments, each simulated for 5 seconds. We provide Table~\ref{tab:training_implementation} which lists the implementation details used for policy optimization, diffusion training, replay-buffer construction, and reward weighting.

\section{Task Reward Designs}
\label{app:task_reward_designs}
In this section, we explain the goal and reward designs of each task used in hierarchical control and zero-shot control.

\subsection{Hierarchical Task}
\paragraph{Steering.}
In the steering task, the goal observation is defined as 
$\vc{g}_t = \tilde{\vc{d}}^{*}_t,$
where $\tilde{\vc{d}}^{*}_t$ denotes the target direction  in the character’s local coordinate frame. The same direction is used as the target facing direction. The task reward is calculated according to
\begin{equation}
r_t^{G}
=
0.7
\exp\left(
-0.25
\left(
v^{*}
-
{\vc{d}^{*}_t}^{T}
\dot{\vc{x}}_{t}^{\mathrm{root}}
\right)^2
\right)
+
0.3
{\vc{d}^{*}_t}^{T}
\vc{h}_{t}^{\mathrm{root}},
\end{equation}
where $\dot{\vc{x}}_{t}^{\mathrm{root}}$ denotes the horizontal root velocity, $\vc{h}_{t}^{\mathrm{root}}$ denotes the root facing direction, and $v^{*}=1.5\,\mathrm{m/s}$ is the target speed. The reward encourages the character to travel along the target direction at the desired speed while aligning its facing direction with the target direction.

\paragraph{Backflip.}
The backflip task rewards a high airborne motion while moving in a target horizontal direction. The goal observation is defined as
$
\vc{g}_t =
\left(
\tilde{\vc{d}}^{*}_t,
h^{\mathrm{root}*}
\right),
$
where $\tilde{\vc{d}}^{*}_t$ denotes the target horizontal direction in the character's local coordinate frame and $h^{\mathrm{root}*}$ denotes the target root height. The task reward is
\begin{equation}
r_t^{G}
=
w^{\mathrm{flip}} r_t^{\mathrm{flip}}
+
w^{\mathrm{dir}} r_t^{\mathrm{dir}},
\qquad
r_t^{\mathrm{flip}}
=
r_t^{\mathrm{height}} r_t^{\mathrm{foot}},
\end{equation}
where
\begin{equation}
\begin{aligned}
r_t^{\mathrm{dir}}
&=
\mathrm{clip}\left(
\frac{\Delta \vc{x}_t}
{\left\lVert \Delta \vc{x}_t \right\rVert_2}
\cdot
\frac{\vc{d}^{*}}
{\left\lVert \vc{d}^{*} \right\rVert_2},
0,1
\right), \\
r_t^{\mathrm{height}}
&=
\mathrm{clip}\left(
\frac{
\max_{\vc{s}\in\tau_t} h^{\mathrm{root}}(\vc{s})
}{
h^{\mathrm{root}*}
},
0,1
\right), \\
r_t^{\mathrm{foot}}
&=
\min_{j\in\{\mathrm{L},\mathrm{R}\}}
\mathrm{clip}\left(
\frac{
\max_{\vc{s}\in\tau_t} h^{j}(\vc{s})
}{
h^{\mathrm{foot}*}
},
0,1
\right).
\end{aligned}
\end{equation}
Here, $\tau_t$ denotes the state sequence within the evaluation window at timestep $t$, $h^{\mathrm{root}*}$ and $h^{\mathrm{foot}*}$ are the target root and foot heights, $\Delta\vc{x}_t$ is the horizontal root displacement from the task starting position, and $\vc{d}^{*}$ is the target horizontal direction. The height rewards saturate once their respective target heights are reached, while the direction reward measures the alignment between the normalized root displacement and target direction.

\paragraph{Location.}
The goal of the location task is represented as
$\vc{g}_t = \vc{\tilde{x}^{*}}_t,$
where $\vc{\tilde{x}^{*}}_t$ denotes the target location expressed in the character's local frame. The reward combines target-position accuracy and facing direction:
\begin{equation}
r_t^{G}
=
0.7 \,
\exp\left(
-0.25
\left\lVert
\vc{x}_{t,xy}^{\mathrm{root}}-\vc{x}_{xy}^{*}
\right\rVert^2
\right)
+
0.3 \,
\max\left(
0,
{\vc{d}}_t^{T}\vc{f}_t
\right),
\end{equation}
where $\vc{x}_{t,xy}^{\mathrm{root}}$ is the horizontal root position, $\vc{d}_t$ is the normalized horizontal direction from the character to the target, and $\vc{f}_t$ is the character's horizontal facing direction. The first term encourages the character to approach the target location, while the second encourages it to face toward the target.

\paragraph{Dodging}
The goal observation of the dodging task is represented as
$
\vc{g}_t =
\left(
\vc{\tilde{p}}_t^{\mathrm{root}},
\vc{\tilde{q}}_t^{\mathrm{root}},
\vc{\tilde{p}}_t^{\mathrm{key}}
\right).
$
Here, $\vc{\tilde{p}}_t^{\mathrm{root}}$ denotes the opponent root position, $\vc{\tilde{q}}_t^{\mathrm{root}}$ denotes the opponent root orientation, and $\vc{\tilde{p}}_t^{\mathrm{key}}$ denotes the opponent key-joint positions, respectively, all represented in the character's local coordinate frame. The reward is defined as 
\begin{equation}
r^{G} = \sum_{t=0}^{T} \mathbb{I}(t),
\label{eq:reward_avoid_main}
\end{equation}
where $\mathbb{I}(t)=1$ upon successful avoidance, $\mathbb{I}(t)=-0.2$ after a collision has been detected, and $\mathbb{I}(t)=0$ otherwise. The collision state remains active until the next opponent appears.

\begin{table}[t]
\centering
\caption{Training implementation details for DSD.}
\label{tab:training_implementation}
\begin{tabular}{ll}
\toprule
\textbf{Parameter} & \textbf{Value} \\
\midrule
$\pi$ Policy Stepsize & $5 \times 10^{-5}$ \\
$V$ Critic Stepsize & $5 \times 10^{-5}$ \\
$D$ Discriminator Stepsize & $5 \times 10^{-5}$ \\ 
$\epsilon_\theta$ Diffusion Stepsize & $5 \times 10^{-5}$ \\
$\gamma$ Discount factor & $0.99$ \\
GAE ($\lambda$) & 0.95  \\
TD($\lambda$) & 0.95 \\
$\mathrm{dim}(\mathcal{Z})$ Latent Space Dimension & 64 \\
$N$ State History & 10 \\
$\mathcal{R}$ Diffusion Buffer Size & $3 \times 10^6$ \\
$\eta$ Diffusion Buffer Subsampling Factor & $32$ \\
$M$ Diffusion Timesteps & $50$ \\
$\mathbb{K}$ Fixed Reward Timesteps & $\{8, 15, 22\}$ \\
$\kappa$ Encoding Scaling Factor & 1 \\
PPO clip threshold & 0.2 \\
$w^{\mathrm{ME}}$ Marginal Entropy Reward Weight & 0.25 \\
$w^{\mathrm{CE}}$ Conditional Entropy Reward Weight & 0.75 \\
$w^{\mathrm{DSD}}$ DSD reward weight & $ 0.4$ \\
$w^{AMP}$ AMP reward weight & $0.6$ \\
\bottomrule
\end{tabular}
\end{table}

\paragraph{Shield Bash}
The shield bash task records opponent's states as a goal observation 
$
\vc{g}_t =
\left(
\vc{\tilde{p}}_t^{\mathrm{root}},
\vc{\tilde{q}}_t^{\mathrm{root}},
\vc{\tilde{p}}_t^{\mathrm{key}}
\right).
$ The task reward combines character-side shield bash quality and opponent-side impact:
\begin{equation}
r^{\mathrm{task}} =
0.5
r^{\mathrm{ratio}}
r^{\mathrm{hitvel}}
+
0.5
r^{\mathrm{impulse}}
r^{\mathrm{veldiff}}.
\end{equation}
The first product encourages shield-led impact: $r^{\mathrm{ratio}}$ makes shield contact dominate the agent contact force, and $r^{\mathrm{hitvel}}$ rewards high pre-impact shield speed. The second product rewards strong opponent impact through $r^{\mathrm{impulse}}$ and deviation from the opponent's nominal running motion through $r^{\mathrm{veldiff}}$. These rewards are computed as
\begin{equation}
\begin{aligned}
r^{\mathrm{ratio}}
&=
\frac{
\left\lVert
\vc{f}^{\mathrm{shield}}
\right\rVert
}{
\sum_{b \in \mathcal{B}}
\left\lVert
\vc{f}^{b}
\right\rVert} \\
r^{\mathrm{hitvel}} &=
\mathrm{clip}\left(
0.7
\left\lVert
\vc{v}^{\mathrm{pre}, \mathrm{shield}}
\right\rVert^2,
0, 1
\right), \\
r^{\mathrm{impulse}} &=
\mathrm{clip}\left(
0.005 \, I,
0, 1
\right), \\
r^{\mathrm{veldiff}} &=
\mathrm{clip}\left(
0.2
\left\lVert
\vc{v}^{\mathrm{curr}, \mathrm{opp}} -
\vc{v}^{\mathrm{ref}, \mathrm{opp}}
\right\rVert^2,
0, 1
\right),
\end{aligned}
\end{equation}
where $\vc{f}^{\mathrm{shield}}$ is the contact force applied to the shield joint, $\vc{f}^{b}$ is the contact force applied to body part $b$, and $\mathcal{B}$ is the set of agent body parts. The terms $\vc{v}^{\mathrm{pre}, \mathrm{shield}}$, $I$, $\vc{v}^{\mathrm{curr}, \mathrm{opp}}$, and $\vc{v}^{\mathrm{ref}, \mathrm{opp}}$ denote the pre-impact shield velocity, the impulse applied to the opponent, the current horizontal opponent velocity, and the reference horizontal opponent velocity, respectively.

\paragraph{Beat Saber Task}
The Beat Saber task goal observation is represented as
\begin{equation}
\vc{g}_t =
\left(
\vc{\tilde{p}}_t^{\mathrm{box}},
\vc{\tilde{v}}_t^{\mathrm{box}},
\vc{\tilde{p}}_t^{\mathrm{est}},
\vc{\tilde{p}}_t^{\mathrm{sword}},
\vc{\tilde{p}}_t^{\mathrm{hand}},
\eta_t^{\mathrm{arrive}},
\eta_t^{\mathrm{end}}
\right).
\end{equation}
All position and velocity quantities are expressed in the heading-aligned local coordinate frame of the character. The box observation consists of the relative box position $\vc{\tilde{p}}_t^{\mathrm{box}}$, relative box velocity $\vc{\tilde{v}}_t^{\mathrm{box}}$, and estimated box position $\vc{\tilde{p}}_t^{\mathrm{est}}$ at the predefined sword-hit location in front of the character. The sword observation consists of the relative sword-tip position $\vc{\tilde{p}}_t^{\mathrm{sword}}$ and relative hand position $\vc{\tilde{p}}_t^{\mathrm{hand}}$. The timing variables $\eta_t^{\mathrm{arrive}}$ and $\eta_t^{\mathrm{end}}$ specify the time until the box reaches the active hit window and the time until it exits the window. The task reward is phase-dependent:
\begin{equation}
r^{G} =
\sum_{t=0}^{T}
\sum_{p \in \mathcal{P}}
m_{p,t} r^p_t,
\quad
\mathcal{P} =
\{\mathrm{Recover}, \mathrm{Slash}, \mathrm{Hit}, \mathrm{Miss}\},
\end{equation}
where $m_{p,t}$ is the binary mask for phase $p$. The phase rewards are defined as 
\begin{equation}
\begin{aligned}
r^{\mathrm{recover}}_t
&=
\mathrm{clip}\left(
10 (d_t - d_{t-1}),
0, 1
\right), \\
r^{\mathrm{slash}}_t
&=
\mathrm{clip}\left(
15 (d_{t-1} - d_t),
0, 1
\right), \\
r^{\mathrm{hit}}_t
&=
\mathrm{clip}\left(
0.3 v_t^{\mathrm{hit}},
0, 1
\right), \\
r^{\mathrm{miss}}_t
&=
\exp\left(
-2 t_t^{\mathrm{diff}}
\right).
\end{aligned}
\end{equation}
Here, $d_t$ denotes the displacement from the sword tip to the estimated hit region, $t_t^{\mathrm{diff}}$ denotes the temporal deviation after a missed strike, and $v_t^{\mathrm{hit}}$ denotes the sword speed at impact. The four phase rewards encourage the sword to move away from the box during recovery, execute a slash toward the box, produce a fast sword impact,  and reduce timing error when the strike misses the box, respectively.

\subsection{Zero-shot Tasks}
\paragraph{Strafe.}
The strafe task evaluates direction-conditioned locomotion control from the offline trajectory pool, where the selected trajectory should travel along one direction while facing another. The task goal is represented as
$ \vc{g} =
\left(
\vc{d}^{*},
\vc{h}^{*}
\right), $
where $\vc{d}^{*}$ denotes the target travel direction and $\vc{h}^{*}$ denotes the target facing direction, both defined on the horizontal plane. For a candidate trajectory $\tau^i$, the Sum fitness function assigns high values to trajectories that move along $\vc{d}^{*}$ at the target speed $v^{*}$ while aligning the root facing direction with $\vc{h}^{*}$:
\begin{equation}
F^{\mathrm{sum}}(\tau^i, \vc{g})
=
\sum_{t=0}^{T}
\left[
0.7
\exp\left(
-0.25
\left(
v^{*}
-
{\vc{d}^{*}}^{T}
\dot{\vc{x}}_{t}^{i}
\right)^2
\right)
+
0.3
{\vc{h}^{*}}^{T}
\vc{h}_{t}^{i}
\right].
\end{equation}
Here, $\dot{\vc{x}}_{t}^{i}$ denotes the horizontal root velocity, $\vc{h}_{t}^{i}$ denotes the root facing direction, and $v^{*}=1.2\,\mathrm{m/s}$ is the target speed.

\paragraph{Run.}
The run task evaluates whether zero-shot control can reach a distant target location with a short arrival time. The target root location $\vc{g} = \vc{x}^{*}$ is sampled from $3\,\mathrm{m}$ to $5\,\mathrm{m}$ around the initial character position. A trajectory is considered successful if its root position enters a target radius $\epsilon$. The Max fitness function assigns higher values to trajectories that enter the target radius earlier, while unsuccessful trajectories receive zero fitness:
\begin{equation}
F^{\mathrm{max}}(\tau^i, \vc{x}^{*})
=
\max_{0 \leq t \leq T}
m_t^i
\exp\left(
-
\mathrm{clip} \left(
t\Delta t - t^{\min}, 0
\right)
\right),
\end{equation}
where $m_t^i$ is a binary mask indicating whether the root position of trajectory $\tau^i$ is within the target radius around $\vc{x}^*$ at timestep $t$, $\Delta t = 0.03$ is the timestep duration, and $t^{\min}=0.5$ denotes the minimum expected arrival time.

\paragraph{Location.}
The location task evaluates whether zero-shot selection can perform target-reaching control using the offline trajectory pool. The task objective is specified by a target root location $\vc{g} = \vc{x}^{*}$ on the horizontal plane. For each candidate trajectory $\tau^i$, we compute the Sum fitness function as 
\begin{equation}
F^{\mathrm{sum}}(\tau^i, \vc{x}^{*})
=
\sum_{t=0}^{T}
\exp \left(
-0.5
\left\lVert
\vc{x}_{t}^{i,\mathrm{root}} - \vc{x}^{*}
\right\rVert_2^2
\right),
\end{equation}
where $\vc{x}_{t}^{i,\mathrm{root}}$ denotes the horizontal root position at timestep $t$.

\paragraph{Reach.}
The reach task evaluates whether zero-shot selection can identify a motion that places the sword tip near a target location. The task objective is specified by a target sword-tip location $\vc{g} = \vc{x}^{*}$. For each candidate trajectory $\tau^i$, we compute the Max fitness function as
\begin{equation}
F^{\mathrm{max}}(\tau^i, \vc{x}^{*})
=
\max_{t}
\exp \left(
-5
\left\lVert
\vc{x}_{t}^{i,\mathrm{sword}} - \vc{x}^{*}
\right\rVert_2^2
\right),
\end{equation}
where $\vc{x}_{t}^{i,\mathrm{sword}}$ denotes the position of the sword tip at timestep $t$. The target is sampled within $1\,\mathrm{m}$ of the character's root.

\paragraph{Jump.}
The jump task evaluates the character to match a target jump height. The goal is defined as a desired root height $\vc{g} = h^{*}$. For a candidate trajectory $\tau^i$, the Max fitness function favors trajectories that reach this height at some timestep:
\begin{equation}
F^{\mathrm{max}}(\tau^i, h^{*})
=
\max_{0 \leq t \leq T}
\exp \left(
-10
\left(
h_{t}^{i,\mathrm{root}} - h^{*}
\right)^2
\right),
\end{equation}
where $h_{t}^{i,\mathrm{root}}$ denotes the root height at timestep $t$.

\paragraph{Duck.}
The duck task evaluates whether zero-shot selection can identify a motion that lowers the character's head and subsequently recovers toward an upright posture. For each candidate trajectory $\tau^i$, we compute the Max fitness function as
\begin{equation}
F^{\mathrm{max}}(\tau^i)
=
\max_{0 \leq t \leq T}
f_{\mathrm{low},t}
\left[
(1-w_{\mathrm{rec}})f_{\mathrm{drop},t}
+
w_{\mathrm{rec}}f_{\mathrm{rec},t}
\right],
\end{equation}
where $f_{\mathrm{low},t}$ evaluates whether the head reaches a sufficiently low height, $f_{\mathrm{drop},t}$ measures the vertical drop from the initial head height, and $f_{\mathrm{rec},t}$ measures the subsequent recovery from the minimum height. We use $w_{\mathrm{rec}}=0.75$ to emphasize recovery rather than simply maintaining a crouched posture.

\paragraph{Dodge.}
The dodge task evaluates whether zero-shot selection can identify a motion that retreats away from an enemy while avoiding excessive lateral, forward, or backward displacement. For each candidate trajectory $\tau^i$, we compute the Sum fitness function as
\begin{equation}
F^{\mathrm{sum}}(\tau^i)
=
\sum_{t=0}^{T}
f_{\mathrm{ret},t}
f_{\mathrm{dir},t}
\left[
(1-w_{\mathrm{face}})
+
w_{\mathrm{face}}f_{\mathrm{face},t}
\right],
\end{equation}
where $f_{\mathrm{ret},t}$ rewards retreat away from the enemy, $f_{\mathrm{dir},t}$ penalizes lateral and forward motion, and $f_{\mathrm{face},t}$ favors facing the enemy. We use $w_{\mathrm{face}}=0.3$, making retreat the primary objective.

\paragraph{Combats.}
The combat tasks evaluate whether zero-shot control enables the character to strike a target opponent with a specified weapon or body part. The tasks include sword attack, shield attack, kick attack, and combo attack. The task goal is represented as $\vc{g} = \vc{x}^{\mathrm{opp}}$,
where $\vc{x}^{\mathrm{opp}}$ denotes the target opponent location. The opponent is approximated by a vertical cylinder with diameter $0.2\,\mathrm{m}$ centered at this location. For each attack task, a target joint $\mathrm{j}$ is selected: the sword for sword attack, the shield for shield attack, and the right foot for kick attack. The Sum fitness function is defined as
\begin{equation}
F^{\mathrm{sum}}(\tau^i, \vc{g})
=
\sum_{t=0}^{T}
m_t^i
\left(
1 -
\exp\left(
-0.1
\left\lVert
\dot{\vc{x}}_{t}^{i,\mathrm{j}}
\right\rVert_2
\right)
\right),
\end{equation}
where $\dot{\vc{x}}_{t}^{i,\mathrm{j}}$ denotes the velocity of the target joint, and $m_t^i$ is a binary mask that is active only when the target joint crosses into the opponent cylinder.

\section{Downstream Behavior Analysis}
\label{app:behavior_analysis}

\paragraph{Backflip Success.}
We use a simple kinematic criterion to determine whether a pretrained policy produces backflip-like motions. In MimicKit, backflips are the only reference motions in which both feet rise to approximately head height, making foot height a practical proxy for identifying the behavior. For each frame in a 10-second evaluation rollout, we consider the corresponding 2-second window and check whether both feet simultaneously reach a height of at least $1.5\,\mathrm{m}$ at any point within that window. A rollout is considered successful if this condition is satisfied for at least $50\%$ of its frames. A pretrained policy is considered to provide the backflip behavior if at least $70\%$ of its 1{,}024 evaluation rollouts are successful.

\paragraph{Shield Bash Analysis.}
Successful shield-bash behavior is identified by the change in enemy velocity produced by shield strikes. For each enemy struck during a rollout, the maximum change in velocity is recorded, and the strike is classified as successful when this value is at least $4.0\,\mathrm{m/s}$. A rollout is considered successful when successful strikes are produced against at least half of the enemies struck during the rollout, rounding up for an odd number of enemies. For example, at least three successful strikes are required when six enemies are struck, and at least four are required when seven enemies are struck. A pretrained policy is considered successful when at least $70\%$ of the 1{,}024 evaluation rollouts satisfy this criterion. The number of independently trained policies satisfying this criterion is reported for each method.

\end{document}